\documentclass{CVM}

\newcommand{\degree}[1]{°\ }
\DeclareUnicodeCharacter{2218}{\degree}
\usepackage{booktabs}
\usepackage{multirow}
\usepackage{xcolor,colortbl}
\newcommand{\gray}{\cellcolor{gray!20}}
\usepackage{amsmath}
\usepackage{graphicx}

\CVMsetup{
type      = {Research/Review Article},
doi       = {s41095-0xx-xxxx-x},
title     = {Multi-task Geometric Estimation of Depth and Surface Normal from Monocular 360° Images},
author    = {Kun Huang$^{1}$, Fang-Lue Zhang$^{1}$\cor{}, Fangfang Zhang$^{1}$, Yu-Kun Lai$^{2}$, Paul L. Rosin$^{2}$, and Neil A. Dodgson$^{1}$\\
\note{}},
runauthor = {K. Huang, F.-L. Zhang, F. Zhang, Y.-K. Lai, P. Rosin, N.A. Dodgson},
abstract  = {
Geometric estimation is required for scene understanding and analysis in panoramic 360° images. Current methods usually predict a single feature, such as depth or surface normal. These methods can lack robustness, especially when dealing with intricate textures or complex object surfaces. We introduce a novel multi-task learning (MTL) network that simultaneously estimates depth and surface normals from 360° images. Our first innovation is our MTL architecture, which enhances predictions for both tasks by integrating geometric information from depth and surface normal estimation, enabling a deeper understanding of 3D scene structure. Another innovation is our fusion module, which bridges the two tasks, allowing the network to learn shared representations that improve accuracy and robustness.
Experimental results demonstrate that our MTL architecture significantly outperforms state-of-the-art methods in both depth and surface normal estimation, showing superior performance in complex and diverse scenes. Our model's effectiveness and generalizability, particularly in handling intricate surface textures, establish it as a new benchmark in 360° image geometric estimation. The code and model are available at \url{https://github.com/huangkun101230/360MTLGeometricEstimation}.

},
keywords  = {360°, depth estimation, surface normal estimation, multi-task learning},
copyright = {The Author(s)},
}

\begin{document}

\maketitle

    \begin{figure}[b] \vskip -4mm
    \small\renewcommand\arraystretch{1.3}
        \begin{tabular}{p{80.5mm}} \toprule\\ \end{tabular}
        \vskip -4.5mm \noindent \setlength{\tabcolsep}{1pt}
        \begin{tabular}{p{3.5mm}p{80mm}}
    $1\quad $ & School of Engineering and Computer Science, Victoria University of Wellington, Wellington, New Zealand. E-mail: Kun Huang, kun.huang@vuw.ac.nz; Fanglue Zhang, fanglue.zhang@vuw.ac.nz; Fangfang Zhang, fangfang.zhang@vuw.ac.nz; Neil Dodgson, neil.dodgson@vuw.ac.nz. Fang-Lue Zhang is the corresponding author.\\
    $2\quad $ & School of Computer Science and Informatics, Cardiff University, Cardiff, UK. Email: Rosin, RosinPL@cardiff.ac.uk; Y.-K. Lai, Yukun.Lai@cs.cardiff.ac.uk.\\
&\hspace{-5mm} Manuscript received: 2022-01-01; accepted: 2022-01-01\vspace{-2mm}
    \end{tabular} \vspace {-3mm}
    \end{figure}

\section{Introduction}\label{sec:introduction}

\begin{figure*}[t]
    \centering
    \includegraphics[width=1.0\textwidth]{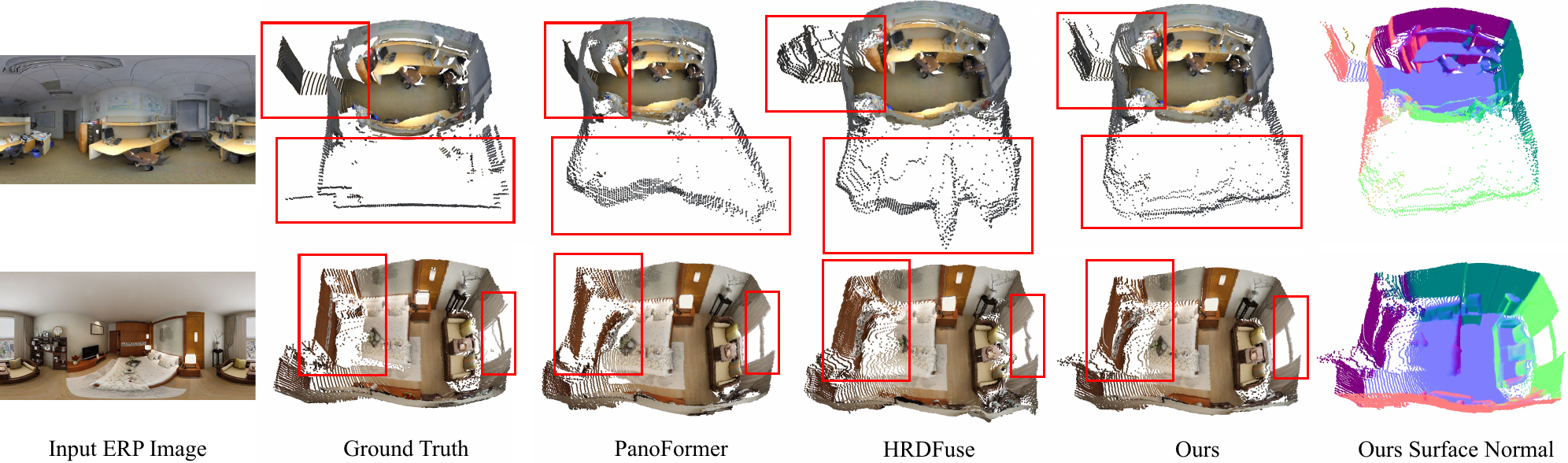}
    \caption{
    Our MTL model provides more accurate geometric estimations for 360° images compared to other methods, particularly in the red rectangle highlighted regions. The results are visualized as 3D point clouds, with both RGB data and color-coded surface normal maps.
    }
    \label{fig:teaser}
\end{figure*}

Multi-task learning (MTL) has emerged as a powerful approach to computer vision. By simultaneously tackling inherently related tasks, MTL leverages shared representations to enhance overall performance, robustness, and generalization across all tasks~\cite{Liu_2019_CVPR,pmlr-v119-standley20a,zhang2024multi,
zhang2022multitask}. We apply MTL to monocular 360° images, simultaneously predicting depth and surface normals.
The 360° depth estimation provides holistic scene information that covers the 360° × 180° field of view (FoV), while 360° surface normal estimation gives insights into the orientation of surfaces within the scene~\cite{long2024adaptive}.
When these tasks are learned together, the model can develop a more comprehensive understanding of the scene's 3D structure, as each task reinforces the other. For instance, accurate depth estimation can inform surface normal prediction by providing context about the relative positioning of objects, while precise surface normal estimation can refine depth predictions by offering additional geometric cues. This synergy between tasks not only enhances the overall accuracy of the model but also improves its ability to generalize to new environments, making MTL for depth and surface normal estimation an important strategy in advancing state-of-the-art computer vision systems, such as those used in indoor perception \cite{medeiros2021promoting} or navigation for cleaning robots. By jointly estimating depth and surface normals, such robots can more effectively understand object distances and surface orientations, enabling them to navigate complex environments efficiently and safely. 360° approaches \cite{huang2023360,li2022deep,rhee2017mr360} can further enhance system capabilities by leveraging predicted depth and surface normal information, enabling dynamic adjustments to varying scene depths and orientations. This flexibility improves spatial alignment and enriches the system's geometric understanding across diverse settings.

Conventional depth estimation methods that rely on perspective projection images struggle with geometric distortions introduced by mapping the entire scene onto the equirectangular projection (ERP) images, which is the most commonly used format for storing and displaying 360° imagery. These distortions, most severe along the vertical axis and intensifying towards the poles, make it challenging for traditional perspective methods to effectively extract features directly from the ERP domain. 
Previous methods~\cite{liao2019spherical, coors2018spherenet} address this problem by extracting reliable features with distortion-aware convolutions and spherical kernels but add computational complexity and often miss global spatial relationships due to the localized nature of convolutional kernels. Alternative approaches~\cite{wang2020bifuse, wang2022bifuse++, jiang2021unifuse} mitigate distortion through different projections, while other models~\cite{li2022omnifusion, shen2022panoformer, ai2023hrdfuse} rely on Vision Transformers (ViTs) to address projection discrepancies and interpret scenes using patch-wise information.
However, these methods often struggle to accurately discern coherent surface regions, particularly in areas that are severely distorted by the spherical projection near the poles of the ERP image. This difficulty is exacerbated by subtle texture variations or repeated patterns, which existing depth estimation techniques struggle to capture accurately. Consequently, depth maps often smooth over these variations, leading to a loss of critical geometric details. While surface normal maps can offer supplementary information to help with these challenges, the integration of depth and surface normal estimation tasks has been less explored in the 360° domain. A multi-task learning approach, where depth and surface normal predictions support and refine each other, can improve both tasks. This approach allows for more accurate depth maps that retain essential geometric details and surface normal maps that are more precise, ultimately enhancing the model’s overall geometric understanding of the 3D scene.

This paper introduces a novel end-to-end deep architecture that leverages a multi-task learning strategy for monocular 360° depth estimation by simultaneously learning surface normals. Inspired by the findings of Standley et al.~\cite{pmlr-v119-standley20a}, which demonstrate that surface normal estimation enhances other tasks in multi-task learning, our approach integrates depth and surface normal predictions to improve overall accuracy.

The proposed network comprises three key components to address distortion issues and enhance depth and surface normal predictions through task knowledge transfer.
First, a shared feature extractor generates features for both depth and surface normal branches, which are processed by two separate spherical distortion-aware ViT networks to address the spherical distortion challenges inherent in 360° imagery. Second, we introduce a novel fusion mechanism that enables knowledge transfer between the spherical ViTs by integrating feature maps from each task. This fusion enhances scene geometry comprehension and improves depth map predictions. Third, task-specific multi-scale transformer decoders are used to handle long-range dependencies, significantly boosting prediction accuracy across various scales.

By simultaneously learning depth and surface normals, our model achieves a more comprehensive understanding of scene structure and geometry, resulting in improved recognition and interpretation of object shapes and spatial relationships, as shown in Fig.~\ref{fig:teaser}. For instance, our model consistently provides clearer segmentation and more accurate geometric details, even in complex regions. The model excels at capturing fine-grained scene structures, which are essential for accurately interpreting depth and surface normals in challenging environments. The insights provided by surface normals enhance the spatial continuity and geometric details of depth estimations, particularly in the areas highlighted by the red rectangles. Additionally, our spherical ViT networks enrich scene comprehension by offering a detailed view of the 3D structure and object layout in panoramic scenes. Extensive experiments demonstrate that our model significantly outperforms state-of-the-art algorithms in both 360° depth and surface normal estimation, showcasing strong generalization in real-world test cases. The contributions of this paper are summarized as follows:
\begin{itemize}
    \item We introduce a novel monocular 360\degree\ MTL architecture for estimating both depth and surface normals. Our approach outperforms state-of-the-art algorithms in performance for both tasks.
    \item We present a fusion module designed to efficiently merge 360\degree\ features in the context of depth and surface normal learning. This module showcases positive sharing among tasks, leading to enhanced scene structure understanding and improved model generalizability.
    \item We conduct comprehensive experiments to evaluate the generalization and robustness of our method in depth and surface normal estimation against state-of-the-art approaches across diverse scenes and datasets. The results reveal that our method consistently outperforms existing approaches on widely recognized benchmarks while maintaining a similar computation time to single-task methods.
\end{itemize}

\section{Related Work}\label{sec:relatedwork}
\subsection{Monocular 360° Depth Estimation}
Various approaches have been taken to tackle the spherical distortion present in ERP imagery for 360° depth estimation. Some methods~\cite{zhuang2022acdnet,sun2021hohonet,zioulis2018omnidepth} directly take the ERP image as input, employing conventional convolutional filters to perceive the spherical distortion field and using other intrinsic information of the scene, such as the indoor layout structure to model the final depth map. In contrast, Liao et al.~\cite{liao2019spherical} and Coors et al.~\cite{coors2018spherenet} introduced distortion-adapted kernels to enable formal convolution operations on ERP images. However, these methods often demand significant computational power, and their effectiveness remains less explored. More recently, bi-projection has emerged as an increasingly popular approach to addressing distortion challenges. This technique involves projecting the distorted image onto a suitable intermediate representation and then reprojecting it back to the original domain. Approaches such as GLPanoDepth~\cite{bai2024glpanodepth}, BiFuse~\cite{wang2020bifuse}, BiFuse++~\cite{wang2022bifuse++}, and UniFuse~\cite{jiang2021unifuse} incorporate both ERP and CP during neural network training. Specifically, BiFuse proposes fusion at both the encoder and decoder stages, while others share the fused feature only at the encoder stage. Recently, tangent projection (TP) has shown potential to address distortion challenges. This is because the transformed patches under TP have smaller FoVs and less distortion compared to the cube faces. For example, 360MonoDepth~\cite{rey2022360monodepth} directly applies a pre-trained perspective depth estimator to project tangent patches and fuses them back into the ERP image to obtain the final depth map. OmniFusion~\cite{li2022omnifusion}, PanoFormer~\cite{shen2022panoformer} and HRDFuse~\cite{ai2023hrdfuse} apply transformer-based architectures to embed geometry information from tangent patches for depth estimation. Recently, Elite360D~\cite{ai2024elite360d} introduced the use of icosahedron projection to enhance geometric information, while Liu et al.~\cite{liu2024estimating} employ a teacher-student architecture to generate comprehensive features for depth estimation. However, approaches that focus solely on depth estimation can result in models that are less robust and have a limited understanding of scene structure. Single-task learning risks overfitting to specific details and missing crucial information about surface orientation and spatial relationships. 

\subsection{Monocular 360° Surface Normal Estimation}
While surface normal estimation has been extensively studied for perspective images, directly applying these methods~\cite{long2024adaptive,long2021adaptive,bae2021estimating, Chen_2023_CVPR,do2020surface} to the 360° domain often yields unsatisfactory results due to spherical distortions. Although 360° surface normal estimation can provide comprehensive information to enhance geometric awareness, it has been less explored compared to 360° depth estimation. Karakottas et al.~\cite{karakottas2019360} introduced HyperSphere, a state-of-the-art method for estimating surface normals in the 360° domain. This approach uses a quaternion loss for supervising surface normal predictions within a CNN architecture. However, their experiments did not cover the widely used datasets in the 360° domain, limiting applicability and preventing it from establishing a standard similar to that of 360° depth estimation. Additionally, the CNN model struggles to efficiently extract features from ERP imagery, and relying solely on surface normal supervision can make predictions sensitive to subtle texture or color changes—obstacles that can be mitigated by incorporating depth information.

\begin{figure*}[t]
    \centering
    \includegraphics[width=1.0\textwidth]{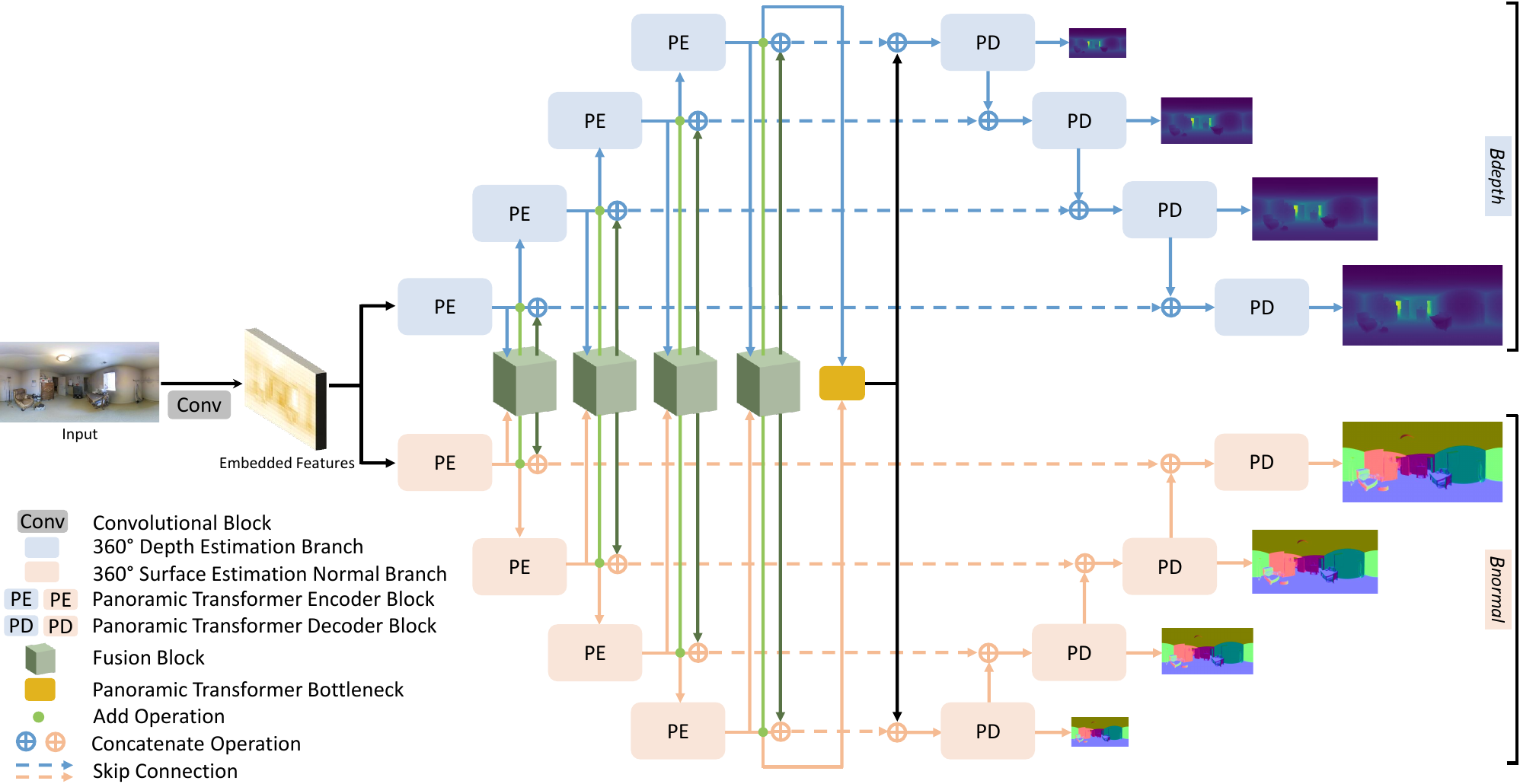}
    \caption{Our network architecture. Our network consists of two branches: $B_{depth}$ (in blue) and $B_{normal}$ (in red), dedicated to depth and surface normal estimation, respectively. A fusion module (in green) is employed to fuse the feature maps between each encoder level of $B_{depth}$ and $B_{normal}$ and feed the fused features into the next encoder level. The fused features are also concatenated with the original depth or normal features and fed to the corresponding decoder blocks. The final depth and normal maps are predicted in a multi-scale manner. 
    }
    \label{fig:network}
\end{figure*}
\subsection{Multi-task Learning for Image Regression} 
Multi-task learning is a form of transfer learning, that addresses multiple tasks simultaneously by leveraging shared domain knowledge across complementary tasks~\cite{pmlr-v119-standley20a, zhang2022task}. For image regression tasks in computer vision, numerous methods~\cite{guizilini2022learning,zhao2022jperceiver,mayer2016large,kendall2018multi} adopt a multi-task strategy to concurrently predict various related tasks, including depth~\cite{bhanushali2024cross}, optical flow~\cite{saxena2024surprising}, scene flow~\cite{liu2023learning}, semantic segmentation~\cite{liu2021pano}, and others~\cite{Dong_2024_CVPR,xu2018pad}, yielding promising results. Recent works have explored different sharing methods for effective knowledge transfer within neural networks, either by manipulating hidden layers or dynamically balancing losses during back-propagation. Approaches such as hard parameter-sharing~\cite{wang2022semi,liu2021pano,kendall2018multi} involve a pipeline with a single encoder and multiple decoders for each task. In contrast, soft parameter-sharing~\cite{zhao2022jperceiver,kundu2019adapt} uses multiple network columns for each task, defining a strategy to share features between columns. Sun et al.~\cite{sun2020adashare} introduce an adaptive method for learning the sharing pattern in multi-task networks, employing a task-specific policy for separate execution paths within a single neural network while still using standard back-propagation. Conversely, others~\cite{chen2018gradnorm,kendall2018multi} have explored an adaptive approach to guide the weight of the loss during back-propagation in MTL. With the established potential and effectiveness of ViT in 360° vision tasks, and the promise of MTL to improve performance through shared representations, our motivation is to explore a multi-task ViT architecture specifically for ERP imagery. By concurrently estimating depth and surface normals, this architecture employs a soft parameter-sharing strategy, which improves robustness and adaptability across diverse scenarios, leading to more accurate 360° depth and surface normal estimation.

\section{Methodology}\label{sec:method}
\subsection{Architecture Overview}
Our MTL architecture leverages the learned representations from both depth and surface normal estimation; this enhances overall scene understanding and improves the accuracy of both tasks.
Depth estimation provides crucial spatial information, while surface normals contribute detailed insights into object perception and surface orientations within a scene. The interaction between these complementary tasks is facilitated by a specially designed fusion block, which promotes seamless integration of information from both tasks. This bidirectional enhancement allows the model to achieve superior performance in both depth and surface normal estimation, leading to a more comprehensive and accurate understanding of the scene. To address the challenges posed by spherical distortion in 360° images, we developed a U-shaped MTL architecture that incorporates distortion-aware ViT blocks, built on the foundation of PanoFormer~\cite{shen2022panoformer} and presented as the panoramic transformer encoder and decoder blocks (PE and PD) in Fig.~\ref{fig:network}. Our transformer decoder incorporates a multi-level structure, focusing on spatial interconnections, handling intricate regional details, and fusing contextual information at varying scales. This hierarchical transformer architecture is applied to both depth and surface normal tasks, resulting in two distinct ViT networks. These networks exchange knowledge in the soft-parameter sharing manner~\cite{ruder2017overview} among their encoders through the proposed fusion block with corresponding scales at each level. 

An overview of our proposed network is shown in Fig.~\ref{fig:network}. Our proposed architecture simultaneously learns to predict the depth and surface normal in a hierarchical structure format, comprising a shared convolutional feature embedding block, fusion blocks, bottleneck, and multi-scale spherical encoders and decoders with spherical distortion awareness. The shared convolutional feature embedding block includes $3\times3$ convolutional layers and a $2\times2$ max pooling layer. 
It aims to extract contextual and salient features from input images for both tasks. Furthermore, employing such a down-sampling layer is crucial in enhancing computational efficiency and reducing parameters in our multi-task model.
The down-sampled features are then directed into two branches ($B_{depth}$ and $B_{normal}$) concurrently. Each branch comprises encoders and decoders and is organized into four hierarchical stages that either halve or double the dimensions and resolution size of features. The feature maps of $B_{depth}$ and $B_{normal}$ are fused through our proposed fusion module at each encoding stage. Finally, the decoder leverages concatenated features from corresponding encoders, fusion modules and decoded feature maps to extract multi-scale depth or surface normal features, followed by an up-sampling step to reconstruct the depth and normal maps at full resolution.

\subsection{Panoramic Transformer Block}
The primary issue in processing panoramic images is the distortion introduced by ERP projection, which differs significantly from the distortion found in perspective images due to the non-uniform spatial warping of features. While our focus is on developing an effective MTL architecture, we adopt the PanoFormer block proposed by Shen et al.~\cite{shen2022panoformer} to specifically address this distortion issue. 
Unlike conventional transformer-based methods that sample features linearly from the input, PanoFormer leverages tangent projection to convert ERP images into a set of tangent patches, each centered on a specific point in the image. By focusing on the pixels surrounding each tangent plane's center, this method captures spherical geometric information more effectively, as the tangent patches avoid the distortions typically present in ERP images, allowing for more accurate feature extraction in 360° imagery. Additionally, this encoder block is enhanced by a locally-optimized feed-forward network~\cite{yuan2021incorporating}, which strengthens local feature interactions within each patch, ensuring that fine-grained details are preserved. The encoder also models token flow relationships between the centers of the tangent patches, enabling the network to understand and capture global dependencies across the entire image. This combination of local refinement and global awareness allows the spherical encoder to better handle the complexities of spherical geometry in 360° images.
The representation of the self-attention mechanism is shown as follows:
\begin{equation}
    P(f,\hat{s}) = \sum\limits_{m}W_m\left[\sum\limits_{(q,k)}A_{mqk}  W'_{m}f(\hat{s}_{mqk}+\Delta s_{mqk})\right]
\end{equation}
where the feature representations $f$ undergo the spherical sampling strategy denoted as $\hat{s}$, involving self-attention heads ($m$), individual tokens ($q$), and their neighboring tokens within a tangent patch ($k$). Learnable weights for each head are denoted by $W_m$ and $W'_{m}$, while $A_{mqk}$ indicates the attention weights assigned to each token, and $\Delta s_{mqk}$ represents the learned flow for individual tokens. 

\subsection{Fusing Depth and Surface Normal Features}
\begin{figure}[t]
    \centering
    \includegraphics[width=1.0\linewidth]{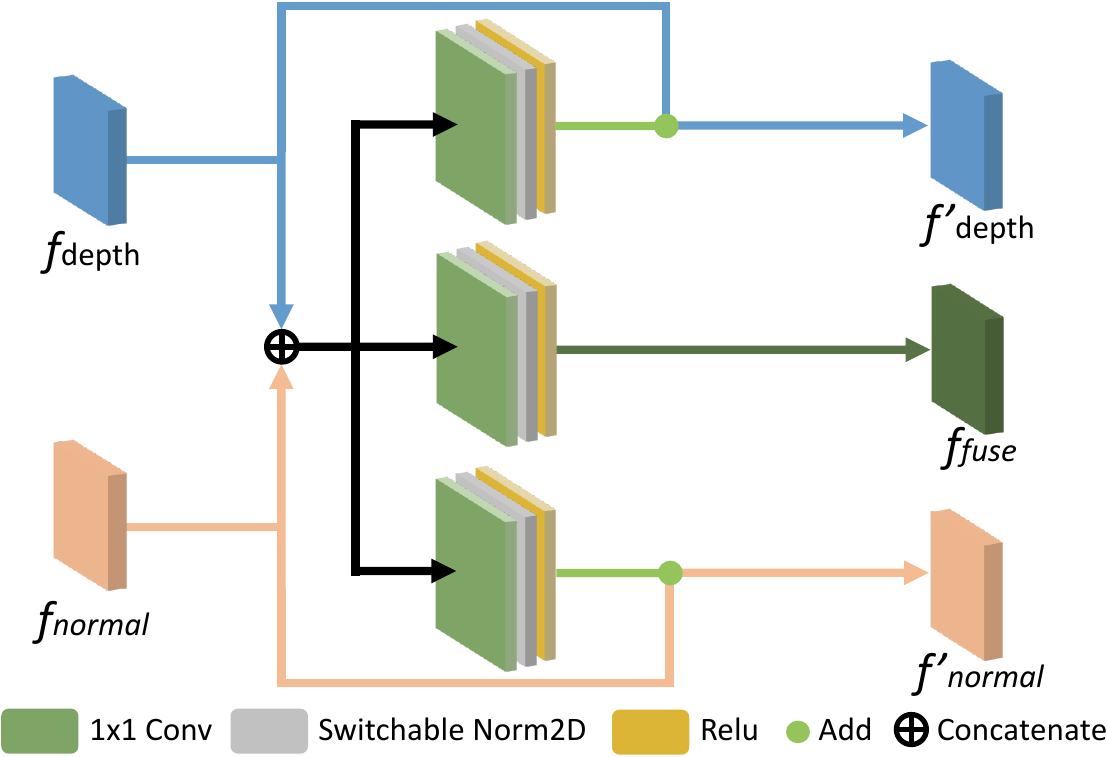}
    \caption{Our proposed fusion module for fusing 360\degree\ depth and surface normal features.
    }
    \label{fig:fusion_block}
\end{figure}

Our fusion module is designed to efficiently extract representations from the two complementary tasks, seamlessly integrated with ViT networks, as depicted in Fig.~\ref{fig:fusion_block}. Our fusion module draws inspiration from BiFuse++. It comprises three blocks, each with an identical structure. Within each block, we employ three convolutional layers: a $1\times1$ convolutional layer, a switchable normalization layer (SwitchableNorm), and a ReLU activation function. Conventional convolutional blocks typically use batch normalization~\cite{ioffe2015batch} to mitigate internal covariate shifts. However, since our network contains both convolutional layers (in fusion blocks) and panoramic ViTs (in main branches), uniformly applying Batch Normalization and Layer Normalization is sub-optimal for our learning task.
Therefore, we adopt Switchable Normalization~\cite{SwitchableNorm} that learns the weights of channel-wise, layer-wise, and batch-wise normalization to adaptively control the behaviour of the normalization layer. By the flexible selection of the most effective normalization strategy for different components and the information learned from different tasks, employing Switchable Normalization improves the generalization and performances of our deep model compared to using a fixed normalization method.

Our fusion module takes the concatenation of depth features ($f_{depth}$) and surface normal features ($f_{normal}$) as input from their respective task branches. This concatenated feature is processed through three individual blocks, each dedicated to learning distinct representations. For the depth block (in blue) and the surface normal block (in red) as shown in Fig.~\ref{fig:fusion_block}, 
each of the two branches predicts a residual map that incorporates additional information from the other task to refine its feature maps. The resulting feature maps ($f'_{depth}$ and $f'_{normal}$) are then propagated to the next encoder level and simultaneously linked to the corresponding decoder block, where they are concatenated with the fused feature map ($f_{fuse}$) from the fusion block. Our fusion module offers several advantages: it facilitates mutual information sharing, allowing each task to benefit from insights gained by the other; it enables the model to adapt to task-specific challenges, focusing on regions where one task provides more reliable information; and it enhances robustness and generalization by reducing sensitivity to noise or inaccuracies in a single task, leading to improved performance across diverse scenes.

\subsection{Multi-scale Spherical Encoding and Decoding}
The proposed ViT decoder ($\hat{P}$) addresses spherical distortion, comprising the same number of blocks as the encoder for each task. It predicts the depth ($\hat{D}_i$) or surface normal map ($\hat{N}_i$) at various scales (denoted by $i$). Each block takes as input the concatenation of the upsampled encoded representations, $\hat{f}_{i}$, which has half the number of channels and double the spatial resolution compared to the previous block. This input is further combined with the task-specific feature and the fused feature through skip-links. After each block, the predictions are produced at various scales using a $3\times3$ convolutional operation followed by a bi-linear interpolation to double the output size. Each task-specific prediction is then passed through a corresponding activation function to constrain the output within its valid range. For instance, Sigmoid activation ($\sigma$) is applied for the depth map to limit values between 0 and 1, while Tanh activation ($\tanh$) is used for the surface normal map to constrain values within the range of [-1, 1]. In our experiments, the model is unable to converge during training without the use of these two activation functions. The process is illustrated as:
\begin{equation}
\hat{D}_i=\sigma\left(\hat{P}_{depth}(\hat{f}_{depth,i} \oplus f'_{depth,i} \oplus f_{fuse, i})\right)
\end{equation}
\vspace{-4mm}
\begin{equation}
\hat{N}_i=\tanh\left(\hat{P}_{normal}(\hat{f}_{normal,i} \oplus f'_{normal,i} \oplus f_{fuse, i})\right)
\end{equation}

Our multi-scale decoder offers a range of advantages. Processing information at various spatial levels of the panoramic scene enables the network to capture both fine details and global context for depth estimation. This adaptability to different object sizes ensures that the network can effectively represent structures of varying scales, promoting robustness to diverse objects. By interpreting the fused feature in a multi-scale manner, a richer representation of the inherent geometry information can be learned, which is valuable for the task of depth estimation. Moreover, the model's generalizability to handle different panoramic scene layouts is enhanced, reducing the risk of over-fitting to common scene structures present in the training data. 

The effectiveness of these introduced components on depth estimation in our MTL network is confirmed through our ablation study (Sec.~\ref{sec:ablation}). 

\subsection{Loss Function}
Our proposed MTL network simultaneously predicts depth and surface normal maps across $S$ scales ($S$ is set to 4 in our experiments), aiming to capture a comprehensive and holistic representation of the scene geometry. During training, we focus on measuring the loss only for valid pixels, and the number of such pixels is denoted as $M$.
The formulated loss functions are shown as follows:
\subsubsection{MSE Loss:} $\mathcal{L}_{Dmse}$ and $\mathcal{L}_{Nmse}$ represent the mean squared error of the estimated map for two tasks, and they are defined as follows:
\begin{equation}
\begin{array}{c}
    \mathcal{L}_{Dmse}=\sum_{j=1}^{M} \| \Delta_D \|_2
\\[0.2cm]
    \mathcal{L}_{Nmse}=\sum_{i=1}^{S}\sum_{j=1}^{M} \| \Delta_\angle \|_2
\end{array}
\end{equation}
where $\Delta_D=\hat{D}_{j}-D_{j}$, and $\hat{D}_{j}$ represents the predicted depth and $D_{j}$ is the ground truth map at the finest scale and the current valid pixel $j$. $\Delta_\angle = \arccos(\hat{N}_{ij} \cdot N_{ij})$ denotes the angular difference.
\subsubsection{Quaternion Loss:} $\mathcal{L}_{quat}$ \cite{karakottas2019360} measures the angular difference between predicted and ground truth normal maps on a pixel-wise basis:
\begin{equation}
    \mathcal{L}_{quat} = \sum_{i=1}^{S}\sum_{j=1}^{M} \arctan\left(\frac{\|\hat{N}_{ij} \times N_{ij}\|}{\hat{N}_{ij} \cdot N_{ij}}\right)
\end{equation}
\subsubsection{Perceptual Loss:} $\mathcal{L}_{Dperc}$ and $L_{Nperc}$ are applied at the finest scale to augment the generation of intricate details in both depth and surface normal predictions:
\begin{equation}
\begin{array}{c}
    \mathcal{L}_{Dperc}=\sum\limits_{j=1}^{M} l_{feat}^{\phi,k}(\hat{D}_{j}, D_{j})\\[0.2cm]
    \mathcal{L}_{Nperc}=\sum\limits_{j=1}^{M} l_{feat}^{\phi,k}(\hat{N}_{j}, N_{j})
\end{array}
\end{equation}
\begin{equation}
    l_{feat}^{\phi,k}(pred,gt) = \frac{1}{C_{k}M}
     \| \phi_{k}(pred_{j}) - \phi_{k}(gt_{j}) \|_2^{2}
\end{equation}
where C denotes the feature's dimension, $\phi$ represents the pre-trained VGG16 network \cite{simonyan2014very}, and $k$ is the $k$-th layer within the network $\phi$.
\subsubsection{Gradient Loss:} In $\mathcal{L}_{grad}$, $\nabla$ represents the sum of the mean absolute differences between the gradients of the predicted depth map and the ground truth maps, computed using Sobel kernel convolutions. This loss is formulated as follows:
\begin{equation}
    \mathcal{L}_{\text{grad}} = \sum\limits_{s=1}^{S}\frac{1}{M} \sum\limits_{i=1}^M \left( \left| |\nabla D_{\text{s}}^i| - |\nabla \hat{D}_{\text{s}}^i| \right| \right)
\end{equation}

The overall loss $\mathcal{L}_{total}$ function of our network is defined as:
\begin{equation}
\begin{split}
    \mathcal{L}_{total} = \lambda_{Dmse}\mathcal{L}_{Dmse} + \lambda_{grad}\mathcal{L}_{grad} + \lambda_{Dperc}\mathcal{L}_{Dperc} + \\
    \lambda_{Nmse}\mathcal{L}_{Nmse} + \lambda_{quat}\mathcal{L}_{quat} + \lambda_{Nperc}\mathcal{L}_{Nperc}
\end{split}
\end{equation}
We assign weights to the depth terms as follows: $\lambda_{Dmse} = 2.0$, $\lambda_{grad} = 1.0$, and $\lambda_{Dperc} = 0.05$. For surface normal terms, we set $\lambda_{Nmse} = 1.0$, $\lambda_{quat} = 10.0$, and $\lambda_{Nperc} = 0.05$. Through our experiments, we noted that employing the specific combination of investigated loss functions and their corresponding weights consistently led to superior outcomes when compared to alternative combinations. This observation carries significant importance in the context of MTL, as an imbalanced adjustment of loss weights or the use of inappropriate loss functions may result in one task dominating over another. Moreover, such misalignments can even lead to the failure of the model to converge, underscoring the critical role of carefully selecting and tuning loss functions for successful multi-task training.

\begin{table*}[t!]
    \caption{\textbf{360° Depth Quantitative comparisons on five benchmarks}. We evaluate our method against state-of-the-art approaches, highlighting improvements over existing best results for each error and accuracy metric as 'Ours-Improved by'.\\ *The evaluation is performed using the corresponding partitions of the 3D60 dataset.}
    \centering
    \begin{tabular}{c|l|cccc|ccc}
        \toprule
        \multirow{2}{*}{\centering Dataset} & \multirow{2}{*}{Method}
        & \multicolumn{4}{c|}{Error metric $\downarrow$} & \multicolumn{3}{c}{Accuracy metric $\uparrow$}  \\
        \cline{3-9}
        & & \raisebox{-0.5ex}{MAE}  & \raisebox{-0.5ex}{ARE} & \raisebox{-0.5ex}{RMSE} & 
        \raisebox{-0.5ex}{RMSElog}
        & \raisebox{-0.5ex}{$\delta_{D1}$} & \raisebox{-0.5ex}{$\delta_{D2}$}  & \raisebox{-0.5ex}{$\delta_{D3}$}\\
        \midrule
        \midrule

        \multirow{7}{*}{\centering 3D60}
         & UniFuse &0.1611	&0.0720	&0.3012	&0.0464	&94.51	&98.87	&99.64\\
         & PanoFormer&0.1244 &0.0617	&0.2234	&0.0386	&96.65	&99.41	&99.80\\
         & HRDFuse &0.1611	&0.0729	&0.2911	&0.0460	&94.72	&98.92	&99.62\\
         & GLPanoDepth &0.1426	&0.0673	&0.2535	&0.0420	&96.00	&99.30	&99.79\\
         & ASNGeo &0.1782	&0.0837	&0.3305	&0.0525	&93.24	&98.68	&99.59\\
         \cline{2-9}
        & \raisebox{-0.4ex}{Ours} &\raisebox{-0.4ex}{\textbf{0.0962}}	&\raisebox{-0.4ex}{\textbf{0.0465}}	&\raisebox{-0.4ex}{\textbf{0.2050}}	&\raisebox{-0.4ex}{\textbf{0.0325}}	&\raisebox{-0.4ex}{\textbf{97.66}}	&\raisebox{-0.4ex}{\textbf{99.50}}	&\raisebox{-0.4ex}{\textbf{99.83}}\\
        
         & \gray \raisebox{-0.4ex}{Ours-Improved by} &\gray\raisebox{-0.4ex}{\textbf{22.67\%}}	&\gray\raisebox{-0.4ex}{\textbf{24.64\%}}	&\gray\raisebox{-0.4ex}{\textbf{8.24\%}}	&\gray\raisebox{-0.4ex}{\textbf{15.80\%}}	&\gray\raisebox{-0.4ex}{\textbf{1.01}}	&\gray\raisebox{-0.4ex}{\textbf{0.09}}	&\gray\raisebox{-0.4ex}{\textbf{0.03}}\\
        \midrule
        \midrule

        \multirow{7}{*}{\centering Stanford2D3D*}
         & UniFuse &0.1539	&0.0683	&0.2884	&0.0462	&95.08	&99.07	&99.72\\
         &  PanoFormer &0.1099	&0.0537	&0.2043	&0.0363	&97.29	&99.61	&\textbf{99.89}\\
         & HRDFuse &0.1396	&0.0614	&0.2606	&0.0412	&96.39	&99.41	&99.81\\
         & GLPanoDepth &0.1530	&0.0716	&0.2599	&0.0442	&95.89	&99.45	&99.85\\
         & ASNGeo &0.1627	&0.0790	&0.3051	&0.0515	&93.51	&99.01	&99.71\\
        \cline{2-9}
        & \raisebox{-0.4ex}{Ours} &\raisebox{-0.4ex}{\textbf{0.0873}}	&\raisebox{-0.4ex}{\textbf{0.0418}}	&\raisebox{-0.4ex}{\textbf{0.1928}}	&\raisebox{-0.4ex}{\textbf{0.0315}}	&\raisebox{-0.4ex}{\textbf{98.02}}	&\raisebox{-0.4ex}{\textbf{99.67}}	&\raisebox{-0.4ex}{99.88}\\
         & \gray \raisebox{-0.4ex}{Ours-Improved by} &\gray\raisebox{-0.4ex}{\textbf{20.56\%}}	&\gray\raisebox{-0.4ex}{\textbf{22.16\%}}	&\gray\raisebox{-0.4ex}{\textbf{5.63\%}}	&\gray\raisebox{-0.4ex}{\textbf{13.22\%}}	&\gray\raisebox{-0.4ex}{\textbf{0.73}}	&\gray\raisebox{-0.4ex}{\textbf{0.06}}	&\gray\raisebox{-0.4ex}{--0.01}\\
        \midrule
        \midrule
        
        \multirow{7}{*}{\centering Matterport3D*}
         & UniFuse &0.1759	&0.0772	&0.3190	&0.0483	&94.22	&98.96	&99.70\\
         & PanoFormer &0.1348	&0.0657	&0.2360	&0.0399	&96.78	&99.50	&\textbf{99.88}\\
         & HRDFuse &0.1719	&0.0764	&0.3022	&0.0469	&94.93	&99.16	&99.75\\
         & GLPanoDepth &0.1515	&0.0709	&0.2627	&0.0430	&96.12	&99.38	&99.83\\
         & ASNGeo &0.1866	&0.0869	&0.3406	&0.0536	&93.32	&98.71	&99.63\\
        \cline{2-9}
        & \raisebox{-0.4ex}{Ours} &\raisebox{-0.4ex}{\textbf{0.1035}}	&\raisebox{-0.4ex}{\textbf{0.0486}}	&\raisebox{-0.4ex}{\textbf{0.2141}}	&\raisebox{-0.4ex}{\textbf{0.0331}}	&\raisebox{-0.4ex}{\textbf{97.74}}	&\raisebox{-0.4ex}{\textbf{99.59}}	&\raisebox{-0.4ex}{99.87}\\
         & \gray \raisebox{-0.4ex}{Ours-Improved by} &\gray\raisebox{-0.4ex}{\textbf{23.22\%}}	&\gray\raisebox{-0.4ex}{\textbf{26.03\%}}	&\gray\raisebox{-0.4ex}{\textbf{9.28\%}}	&\gray\raisebox{-0.4ex}{\textbf{17.04\%}}	&\gray\raisebox{-0.4ex}{\textbf{0.96}}	&\gray\raisebox{-0.4ex}{\textbf{0.09}}	&\gray\raisebox{-0.4ex}{--0.01}\\
        \midrule
        \midrule

        \multirow{7}{*}{\centering SunCG*}
         & UniFuse &0.1071	&0.0540	&0.2401	&0.0392	&95.19	&98.32	&99.30\\
         & PanoFormer &0.0969	&0.0534	&0.1890	&0.0354	&94.87	&98.83	&99.51\\
         & HRDFuse &0.1366	&0.0690	&0.2744	&0.0465	&92.15	&97.42	&98.89\\
         & GLPanoDepth &0.0957	&0.0486	&0.2094	&0.0361	&95.60	&98.82	&99.54\\
         & ASNGeo &0.1588	&0.0752	&0.3132	&0.0489	&92.63	&98.22	&99.31\\
        \cline{2-9}
        & \raisebox{-0.4ex}{Ours} &\raisebox{-0.4ex}{\textbf{0.0718}}	&\raisebox{-0.4ex}{\textbf{0.0405}}	&\raisebox{-0.4ex}{\textbf{0.1756}}	&\raisebox{-0.4ex}{\textbf{0.0303}}	&\raisebox{-0.4ex}{\textbf{97.14}}	&\raisebox{-0.4ex}{\textbf{99.04}}	&\raisebox{-0.4ex}{\textbf{99.60}}\\
         & \gray \raisebox{-0.4ex}{Ours-Improved by} &\gray\raisebox{-0.4ex}{\textbf{24.97\%}}	&\gray\raisebox{-0.4ex}{\textbf{16.67\%}}	&\gray\raisebox{-0.4ex}{\textbf{7.09\%}}	&\gray\raisebox{-0.4ex}{\textbf{14.41\%}}	&\gray\raisebox{-0.4ex}{\textbf{1.54}}	&\gray\raisebox{-0.4ex}{\textbf{0.21}}	&\gray\raisebox{-0.4ex}{\textbf{0.06}}\\
        \midrule
        \midrule
        \multirow{7}{*}{\centering Structured3D}
         & UniFuse &0.2581	&0.2149	&0.4133	&0.1142	&75.25	&91.09	&95.61\\
         & PanoFormer &0.3097 &0.2697	&0.4804	&0.1283	&71.84	&88.28	&94.06\\
         & HRDFuse &0.3141	&0.3090	&0.4867	&0.1331	&70.72	&87.94	&93.89\\
         & GLPanoDepth &0.5028	&0.4539	&0.6992	&0.1800	&52.66	&76.25	&87.68\\
         & ASNGeo &0.2954	&0.2469	&0.4595	&0.1224	&73.75	&89.87	&94.92\\

        \cline{2-9}
        & \raisebox{-0.4ex}{Ours} &\raisebox{-0.4ex}{\textbf{0.2053}}	&\raisebox{-0.4ex}{\textbf{0.1684}}	&\raisebox{-0.4ex}{\textbf{0.3428}}	&\raisebox{-0.4ex}{\textbf{0.0940}}	&\raisebox{-0.4ex}{\textbf{82.79}}	&\raisebox{-0.4ex}{\textbf{93.63}}	&\raisebox{-0.4ex}{\textbf{96.74}}\\
         & \gray\raisebox{-0.4ex}{Ours-Improved by}  &\gray\raisebox{-0.4ex}{\textbf{20.46\%}}	&\gray\raisebox{-0.4ex}{\textbf{21.64\%}}	&\gray\raisebox{-0.4ex}{\textbf{17.06\%}}	&\gray\raisebox{-0.4ex}{\textbf{17.69\%}}	&\gray\raisebox{-0.4ex}{\textbf{7.54}}	&\gray\raisebox{-0.4ex}{\textbf{2.54}}	&\gray\raisebox{-0.4ex}{\textbf{1.13}}\\

        \midrule
        \midrule
         \multicolumn{2}{c|}{Ours Average Improvement}&\raisebox{-0.4ex}{\textbf{21.57\%}}	&\raisebox{-0.4ex}{\textbf{23.14\%}}	&\raisebox{-0.4ex}{\textbf{12.65\%}}	&\raisebox{-0.4ex}{\textbf{16.75\%}}	&\raisebox{-0.4ex}{\textbf{4.28}}	&\raisebox{-0.4ex}{\textbf{1.32}}	&\raisebox{-0.4ex}{\textbf{0.58}}\\
        \bottomrule
    \end{tabular}
    \label{tab:depth_quantitative_comparison}
\end{table*}

\section{Experiments and Results}\label{sec:results}
We validate our method on five widely recognized panoramic benchmark datasets: 3D60~\cite{zioulis2018omnidepth}, Structured3D~\cite{zheng2020structured3d}, Stanford2D3D~\cite{armeni2017joint}, Matterport3D~\cite{chang2017matterport3d}, and SunCG~\cite{song2017semantic}. Both quantitative and qualitative evaluations are conducted for depth and surface normal estimation tasks, comparing our approach against state-of-the-art methods in both the 360° and perspective domains. Given the limited previous work on MTL models in the 360° domain, our comparisons are primarily against existing single-task learning methods. For 360° depth estimation, we compare our model with GLPanoDepth, PanoFormer, HRDFuse, and UniFuse. In the 360° surface normal estimation task, we compare against the current state-of-the-art, HyperSphere, and adapt the prediction layers of UniFuse, PanoFormer, and OmniFusion to enable surface normal estimation, allowing for a direct comparison of network architectures. There are no prior MTL models designed for 360° imagery; therefore, to provide a comparison against an existing MTL model, we retrain the recently-published perspective-based MTL method ASN\-Geo using 360° data for both tasks. We also conduct an ablation study to assess the key components of our approach, focusing on the depth estimation task, and we further evaluate the computational performance of our method.

\subsection{Evaluation Metric and Datasets}
We assess the performance of depth estimation using four standard error metrics: mean absolute error (MAE), absolute relative error (ARE), root mean square error (RMSE), and logarithmic root mean square error (RMSElog). Additionally, we use three accuracy metrics to evaluate the percentage of pixels where the ratio ($\delta_{D}$) of the difference between the predicted depth map and the ground truth is less than $1.25^1$, $1.25^2$, and $1.25^3$. For surface normal estimation, we use three standard error metrics: mean error, median error, and mean square error (MSE), along with five accuracy metrics that measure the percentage of pixels where the angular difference ($\delta_{N}$) between the predicted normals and the ground truth is less than 5°, 7.5°, 11.25°, 22.5°, and 30°. To ensure fair comparisons, we apply consistent experimental settings across all methods. Detailed specifications for each dataset are provided below.

\subsubsection{3D60 Dataset} 3D60 is a panoramic dataset encompassing RGB, depth, and surface normal data at resolutions of $256 \times 512$, captured in diverse scenes. It includes two real-world indoor scanning environments, Stanford2D3D and Matterport3D, and a synthetic dataset from SunCG, introducing an inherent distribution gap for improved model generalizability. We follow the data split in HyperSphere, as recommended in the official introduction. It is important to note that Matterport3D lacks ground truth for surface normals, and Stanford2D3D's surface normal maps lack consistently aligned vector directions across their data. Additionally, 3D60 faces limitations in rendering imagery that could impact the depth estimation task. As a result, our evaluations focused on specific subsets within the 3D60 dataset.

\subsubsection{Structured3D Dataset} Structured3D constitutes an extensive synthetic dataset, comprising 21,835 panoramic data instances at a resolution of $512 \times 1024$ across 3500 scenes. The dataset includes RGB images illuminated with cold, normal, and warm lighting, along with various annotations, such as depth, surface normal, and semantic segmentation. We preprocess the dataset, forming examples in an 8:1:1 ratio, resulting in 2,181 test data instances with randomly selected lighting conditions.

\subsection{Implementation Details}
Our experiments are conducted using a single CPU core of an Intel Xeon W-2133 along with an RTX 3090 GPU. The batch size is 2, and the input resolution is $256 \times 512$. We employe the Adam optimizer with default settings, initialising the learning rate at $1\times 10^{-4}$ and decreasing by half every 12 epochs. The training process extended to 120 epochs, with early stopping implemented at the 12th epoch if no further improvements are achieved.

\begin{table*}[t!]
    \caption{\textbf{360° Surface Normal Quantitative comparisons on five benchmarks}. We evaluate our method against state-of-the-art approaches, highlighting improvements over existing best results for each error and accuracy metric as 'Ours-Improved by'.\\ *The evaluation is performed using the corresponding partitions of the 3D60 dataset.}
    \centering
        \begin{tabular}{c|l|ccc|ccccc}
            \toprule
            \multirow{2}{*}{\centering Dataset} & \multirow{2}{*}{Method}
            & \multicolumn{3}{c|}{Error metric $\downarrow$} & \multicolumn{5}{c}{Accuracy metric $\uparrow$}  \\
            \cline{3-10}
            & & \raisebox{-0.5ex}{Mean}  & \raisebox{-0.5ex}{Median} & \raisebox{-0.5ex}{MSE}  & \raisebox{-0.5ex}{$\delta_{N1}$} & \raisebox{-0.5ex}{$\delta_{N2}$}  & \raisebox{-0.5ex}{$\delta_{N3}$}  & \raisebox{-0.5ex}{$\delta_{N4}$} & \raisebox{-0.5ex}{$\delta_{N5}$} \\
            \midrule
            \midrule
    
            \multirow{7}{*}{\centering 3D60}
             & UniFuse&6.5829	&0.5169	&268.6102	&76.09	&78.82	&82.46	&89.59	&92.28\\
             & PanoFormer&17.2109	&6.4281	&906.0848	&50.72	&55.15	&60.70	&72.83	&78.10\\
             & OmniFusion&7.7549	&1.3175	&301.8934	&72.35	&76.01	&80.2	&88.22	&91.26\\
             &  HyperSphere&5.6176	&\textbf{0.2421}	&215.0301	&77.34	&79.99	&83.84	&91.11	&93.71\\
             &  ASNGeo&32.8173	&28.0937	&1214.0923	&0.03	&0.04	&0.06	&0.46	&64.11\\
            \cline{2-10}
            & \raisebox{-0.4ex}{Ours} &\raisebox{-0.4ex}{\textbf{5.2394}}	&\raisebox{-0.4ex}{0.3025}	&\raisebox{-0.4ex}{\textbf{187.1651}}	&\raisebox{-0.4ex}{\textbf{78.19}}	&\raisebox{-0.4ex}{\textbf{81.25}}	&\raisebox{-0.4ex}{\textbf{85.12}}	&\raisebox{-0.4ex}{\textbf{91.85}}	&\raisebox{-0.4ex}{\textbf{94.35}}\\
            
             & \gray \raisebox{-0.4ex}{Ours-Improved by} &\gray\raisebox{-0.4ex}{\textbf{6.73\%}}&\gray\raisebox{-0.4ex}{--24.94\%}	&\gray\raisebox{-0.4ex}{\textbf{12.96\%}}	&\gray\raisebox{-0.4ex}{\textbf{0.85}}	&\gray\raisebox{-0.4ex}{\textbf{1.27}}	&\gray\raisebox{-0.4ex}{\textbf{1.28}}	&\gray\raisebox{-0.4ex}{\textbf{0.73}}	&\gray\raisebox{-0.4ex}{\textbf{0.64}}\\
            \midrule
            \midrule
    
            \multirow{7}{*}{\centering Stanford2D3D*}
             & UniFuse&6.9502	&0.4675	&297.777	&76.34	&78.77	&82.21	&88.57	&91.29\\
             & PanoFormer &17.2017	&7.2088	&849.7950	&48.21	&52.98	&59.04	&72.13	&78.08\\
             & OmniFusion &8.0590	&1.2903	&322.0786	&72.42	&76.14	&80.31	&87.46	&90.47\\
             & HyperSphere &6.0463	&\textbf{0.2242}	&244.3175	&77.48	&79.76	&83.26	&89.73	&92.49\\
             &  ASNGeo&33.3921	&28.4364	&1263.7831	&0.04	&0.06	&0.09	&0.39	&60.84\\
            \cline{2-10}
            & \raisebox{-0.4ex}{Ours} &\raisebox{-0.4ex}{\textbf{5.7956}}	&\raisebox{-0.4ex}{0.3222}	&\raisebox{-0.4ex}{\textbf{219.0471}}	&\raisebox{-0.4ex}{\textbf{77.74}}	&\raisebox{-0.4ex}{\textbf{80.27}}	&\raisebox{-0.4ex}{\textbf{83.77}}	&\raisebox{-0.4ex}{\textbf{90.29}}	&\raisebox{-0.4ex}{\textbf{93.02}}\\
             & \gray \raisebox{-0.4ex}{Ours-Improved by} &\gray\raisebox{-0.4ex}{\textbf{4.15\%}}	&\gray\raisebox{-0.4ex}{--43.7\%}	&\gray\raisebox{-0.4ex}{\textbf{10.31\%}}	&\gray\raisebox{-0.4ex}{\textbf{0.26}}	&\gray\raisebox{-0.4ex}{\textbf{0.51}}	&\gray\raisebox{-0.4ex}{\textbf{0.51}}	&\gray\raisebox{-0.4ex}{\textbf{0.56}}	&\gray\raisebox{-0.4ex}{\textbf{0.53}}\\
            \midrule
            \midrule
            
            \multirow{7}{*}{\centering Matterport3D*}
             & UniFuse &7.2675	&0.6434	&289.4016	&72.91	&76.04	&80.22	&88.59	&91.62\\
             & PanoFormer &18.1228	&7.3137	&944.5936	&47.56	&52.20	&58.10	&71.21	&76.84\\
             & OmniFusion &8.500	&1.5493	&327.3145	&69.16	&73.08	&77.75	&87.06	&90.49\\
             &  HyperSphere &6.2324	&\textbf{0.3041}	&231.6290	&74.13	&77.23	&81.70	&90.29	&93.23\\
             &  ASNGeo&33.3434	&28.3574	&1254.0368	&0.03	&0.04	&0.06	&0.45	&61.06\\
            \cline{2-10}
            & \raisebox{-0.4ex}{Ours} &\raisebox{-0.4ex}{\textbf{5.7579}}	&\raisebox{-0.4ex}{0.3677}	&\raisebox{-0.4ex}{\textbf{199.5677}}	&\raisebox{-0.4ex}{\textbf{75.29}}	&\raisebox{-0.4ex}{\textbf{78.92}}	&\raisebox{-0.4ex}{\textbf{83.42}}	&\raisebox{-0.4ex}{\textbf{91.20}}	&\raisebox{-0.4ex}{\textbf{93.99}}\\
             & \gray \raisebox{-0.4ex}{Ours-Improved by} &\gray\raisebox{-0.4ex}{\textbf{7.61\%}}	&\gray\raisebox{-0.4ex}{--20.94\%}	&\gray\raisebox{-0.4ex}{\textbf{13.84\%}}	&\gray\raisebox{-0.4ex}{\textbf{1.16}}	&\gray\raisebox{-0.4ex}{\textbf{1.69}}	&\gray\raisebox{-0.4ex}{\textbf{1.72}}	&\gray\raisebox{-0.4ex}{\textbf{0.91}}	&\gray\raisebox{-0.4ex}{\textbf{0.77}}\\
            \midrule
            \midrule
    
            \multirow{7}{*}{\centering SunCG*}
             & UniFuse &3.3994	&0.0416	&154.5543	&89.01	&90.31	&91.98	&94.75	&95.92\\
             & PanoFormer &13.6272	&2.3037	&805.9224	&65.38	&68.63	&72.32	&79.80	&83.06\\
             & OmniFusion &4.3822	&0.3854	&177.3558	&85.51	&87.96	&90.25	&93.75	&95.22\\
             &  HyperSphere &2.6630	&\textbf{0.0031}	&118.1888	&90.46	&91.61	&93.22	&95.83	&96.86\\
             &  ASNGeo&30.0883	&26.6731	&1001.082	&0	&0	&0	&0.59	&79.91\\
            \cline{2-10}
            & \raisebox{-0.4ex}{Ours}&\raisebox{-0.4ex}{\textbf{2.5345}}	&\raisebox{-0.4ex}{0.0067} &\raisebox{-0.4ex}{\textbf{103.0089}}	&\raisebox{-0.4ex}{\textbf{90.58}}	&\raisebox{-0.4ex}{\textbf{91.84}}	&\raisebox{-0.4ex}{\textbf{93.45}}	&\raisebox{-0.4ex}{\textbf{96.07}}	&\raisebox{-0.4ex}{\textbf{97.11}}\\
             & \gray \raisebox{-0.4ex}{Ours-Improved by} &\gray\raisebox{-0.4ex}{\textbf{4.83\%}}	&\gray\raisebox{-0.4ex}{--116.08\%}	&\gray\raisebox{-0.4ex}{\textbf{12.84\%}}	&\gray\raisebox{-0.4ex}{\textbf{0.12}}	&\gray\raisebox{-0.4ex}{\textbf{0.23}}	&\gray\raisebox{-0.4ex}{\textbf{0.23}}	&\gray\raisebox{-0.4ex}{\textbf{0.24}}	&\gray\raisebox{-0.4ex}{\textbf{0.25}}\\
            \midrule
            \midrule
            
            \multirow{7}{*}{\centering Structured3D}
             & UniFuse &10.4186	&0.7087	&576.2404	&70.99	&76.28	&78.91	&84.11	&86.66\\
             & PanoFormer &20.2634	&8.6808	&1157.807	&47.02	&52.68	&58.12	&68.7	&73.75\\
             & OmniFusion &12.0589	&2.0627	&634.7285	&65.79	&71.9	&75.67	&82.02	&85.04\\
             &  HyperSphere &9.4531	&\textbf{0.2763}	&517.8832	&\textbf{72.76}	&77.73	&79.81	&84.97	&87.62\\
             &  ASNGeo&36.2867	&30.3338	&1538.0035	&0.01	&0.01	&0.01	&0.12	&53.12\\
            \cline{2-10}
            & \raisebox{-0.4ex}{Ours} &\raisebox{-0.4ex}{\textbf{8.9783}}	&\raisebox{-0.4ex}{0.4831}	&\raisebox{-0.4ex}{\textbf{469.0207}}	&\raisebox{-0.4ex}{72.51}	&\raisebox{-0.4ex}{\textbf{77.87}}	&\raisebox{-0.4ex}{\textbf{80.65}}	&\raisebox{-0.4ex}{\textbf{86.02}}	&\raisebox{-0.4ex}{\textbf{88.58}} \\
             & \gray\raisebox{-0.4ex}{Ours-Improved by}  &\gray\raisebox{-0.4ex}{\textbf{5.02\%}}	&\gray\raisebox{-0.4ex}{--74.87\%}	&\gray\raisebox{-0.4ex}{\textbf{9.44\%}}
    &\gray\raisebox{-0.4ex}{--0.25}	&\gray\raisebox{-0.4ex}{\textbf{0.14}}	&\gray\raisebox{-0.4ex}{\textbf{0.84}}	&\gray\raisebox{-0.4ex}{\textbf{1.05}}	&\gray\raisebox{-0.4ex}{\textbf{0.96}} \\
            \midrule
            \midrule
            \multicolumn{2}{c|}{Ours Average Improvement}&{\textbf{5.88\%}}	&{--49.91\%}	&{\textbf{11.20\%}}	&{\textbf{0.30}}	&{\textbf{0.71}}	&{\textbf{1.06}}	&{\textbf{0.89}}	&{\textbf{0.80}} \\
            \bottomrule
        \end{tabular}
    \label{tab:normal_quantitative_comparison}
\end{table*}
\subsection{Experimental Results}

\begin{figure*}[t]
    \centering
    \includegraphics[width=1.0\textwidth]{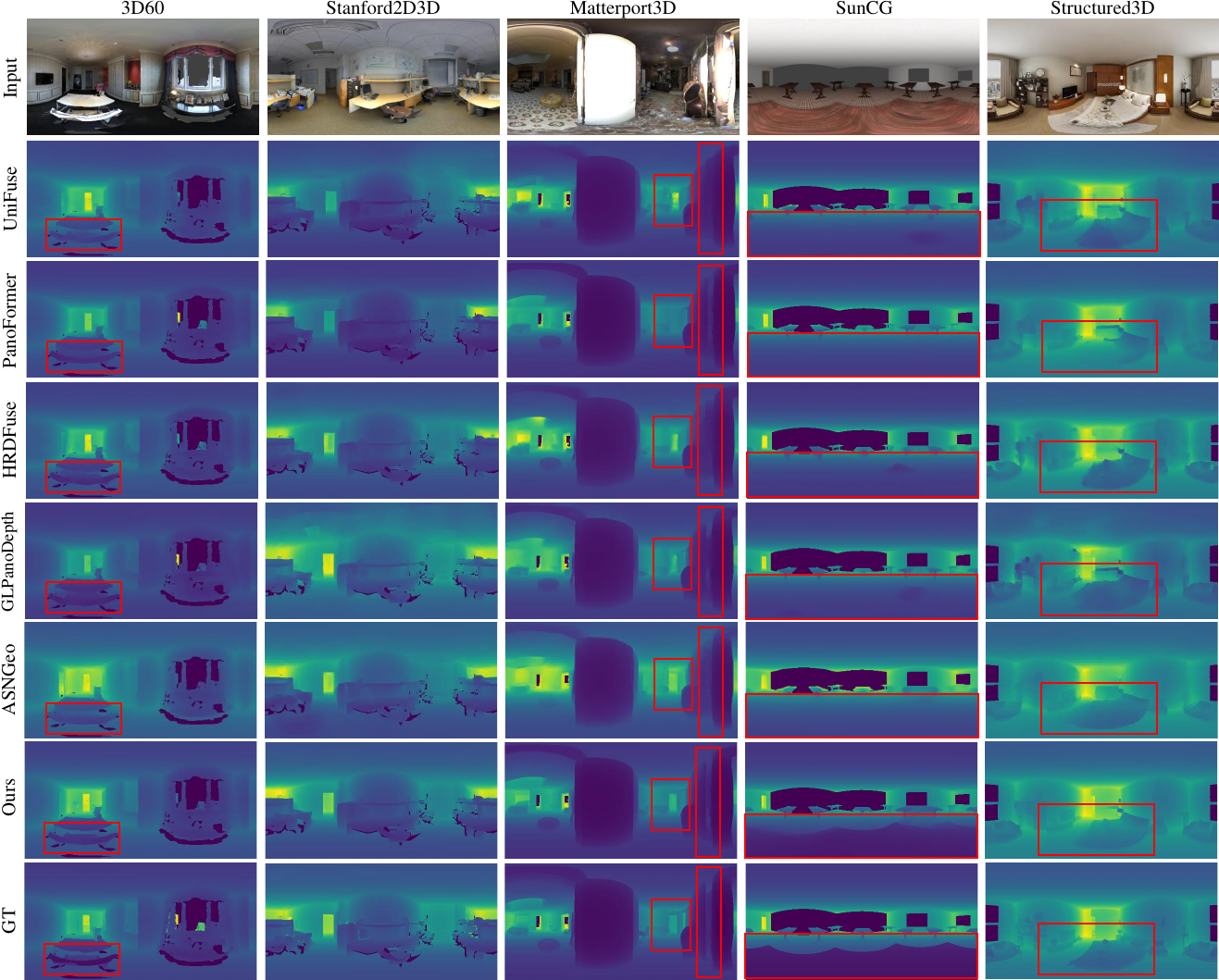}
    \caption{Qualitative 360° depth comparisons were conducted on diverse datasets including 3D60, Stanford2D3D, Matterport3D, SunCG, and Structured3D. The areas outlined in red highlight regions where our approach notably enhances object boundaries, providing a more accurate representation of the overall scene geometry.
    }
    \label{fig:depth_quali}
\end{figure*}

\begin{figure*}[t]
    \centering
    \includegraphics[width=1.0\textwidth]{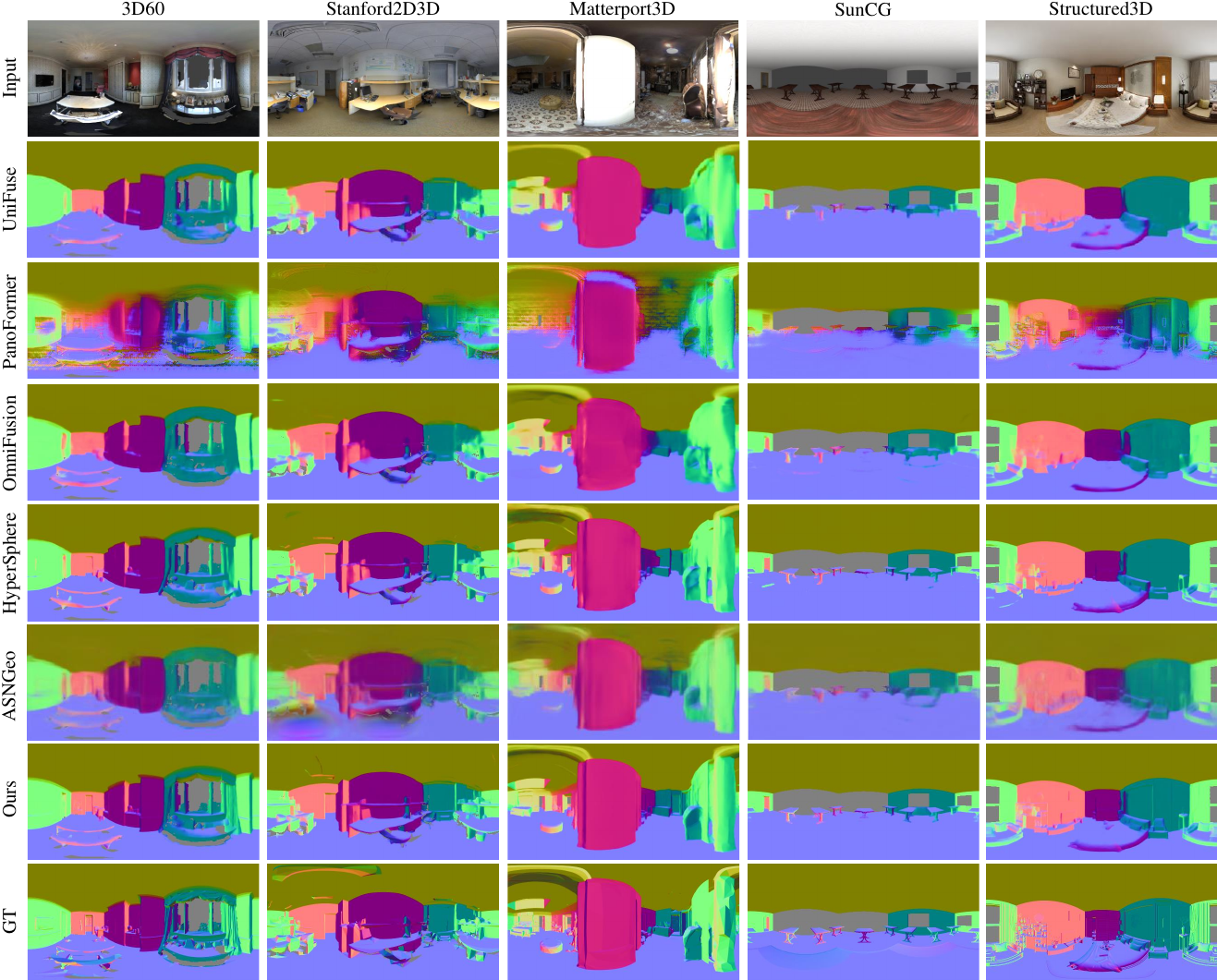}
    \caption{Qualitative 360° surface normal comparisons among HyperSphere, ASNGeo, the adapted UniFuse, PanoFormer, OmniFusion and our method. The best viewing experience in color.
    }
    \label{fig:normal_quali}
\end{figure*}

For both tasks, we conduct a quantitative comparison between our proposed model and state-of-the-art methods across the five datasets, as detailed in Tables~\ref{tab:depth_quantitative_comparison} and \ref{tab:normal_quantitative_comparison}. To ensure a fair evaluation, we retrain all models using their official hyper-parameter settings and identical data splits. Our method consistently outperforms existing approaches in both tasks, setting a new state-of-the-art across all five benchmarks. Specifically, for 360° depth estimation, it demonstrates an average improvement of 21.57\% in MAE, 23.14\% in ARE, 12.65\% in RMSE, and 16.75\% in RMSElog, surpassing previous top-performing depth estimation methods. For 360° surface normal prediction, it achieves a 5.88\% improvement in Mean error and 11.20\% in MSE, although with a higher Median error of 49.91\% on average (see our discussion of limitations in Sec.~\ref{sec:limitation}).

Our model demonstrates a significant performance advantage across all benchmarks for the depth estimation task. Specifically, it achieves substantial improvements of 22.67\%, 24.64\%, and 15.80\% on 3D60; 20.56\%, 22.16\%, and 13.22\% on Stanford2D3D; 23.22\%, 26.03\%, and 17.04\% on Matterport3D; and 24.97\%, 16.67\%, and 14.41\% on SunCG for MAE, ARE, and RMSElog, respectively. These lower error metrics indicate that our model produces more accurate and reliable depth maps, reducing discrepancies between predicted and actual depths. On the surface normal prediction task, our model also shows considerable improvement, with gains of 6.73\% and 12.96\% on 3D60; 4.15\% and 10.31\% on Stanford2D3D; 7.61\% and 13.84\% on Matterport3D; and 4.83\% and 12.84\% on SunCG for the Mean and MSE error metrics, respectively. These improvements indicate that our model delivers more precise surface orientation information, resulting in a deeper understanding of the scene's geometry. This highlights our MTL architecture's effectiveness, which simultaneously enhances both tasks. The lower error metrics for depth and surface normal estimation reflect a more accurate reconstruction of scene depth and finer surface detail recognition. These advances demonstrate the superior performance of our model over state-of-the-art methods, establishing it as a new benchmark in 360° image geometric estimation.

To assess the generalizability of our MTL model, we applied the models trained on the 3D60 dataset directly to the Structured3D dataset, revealing a significant performance gap between our model and others. While UniFuse demonstrates strong generalization in the depth task, our model surpasses it by 20.46\% for MAE, 21.64\% for ARE, and achieves a 7.54 higher accuracy score for depth predictions within a difference of 1.25 with the ground truth. In the surface normal task, our model outperforms HyperSphere, which previously exhibited the best performance. Specifically, our model shows a 5.02\% lower Mean error, and a 9.44\% lower MSE error, but slightly decreased accuracy (0.25) for angular differences with ground truth under 5°. These differences can be attributed to Structured3D's synthetic nature, featuring a unique depth range distribution and a mix of small objects with subtle depth and surface changes, like cups, alongside significant depth variations, such as hollow shelves with books. The limitations of previous methods, which focused exclusively on feature representations learned from the single task, become evident in their reduced generalizability and effectiveness across diverse datasets. Our MTL architecture addresses these challenges, enhancing the adaptability and robustness of our model in various scenarios and across different datasets.

To provide a comparison against another MTL method, we apply a recent MTL method designed for perspective images, ASNGeo, to directly handle 360° data for both depth and surface normal estimation. The quantitative results show that ASNGeo consistently underperforms across all five benchmarks, delivering the worst results compared to other methods. This outcome underscores the limitations of directly applying perspective-based MTL architectures to the 360° domain, where they fail to adequately address the unique challenges posed by spherical distortion and the non-uniform spatial relationships inherent in panoramic imagery. These findings highlight the necessity and effectiveness of our proposed MTL architecture specifically designed for 360° imagery. Unlike perspective-based methods, our approach successfully adapts to the unique geometric properties of panoramic images, resulting in significantly better performance and demonstrating the critical need for specialized solutions in the 360° domain.

We present qualitative comparisons for each task across various methods in Fig.~\ref{fig:depth_quali} and \ref{fig:normal_quali}, showcasing results from a single test instance for each of the five benchmarks. For the depth task, we highlight critical regions in red rectangles to emphasize the differences between our method and existing approaches. As shown in the figures, our model consistently captures finer details, rendering sharper and more complete object boundaries within the scenes. Notably, our model exhibits a superior understanding of the geometric structure, as demonstrated in the Stanford2D3D example, where it accurately captures the scale and proportions of the entire scene with minimal color discrepancies compared to other methods. In the qualitative comparisons for surface normals, we visualize the predicted normal vectors, with invalid areas represented in gray. Our model outperforms state-of-the-art methods by providing more accurate representations of geometric structures and surface orientations, further illustrating its effectiveness in enhancing scene understanding and geometry perception.

\subsection{Discussion of Computational Performance}
To investigate the extra computational cost of using a multi-task approach, we compare computation times between our model and PanoFormer on an RTX3090. As detailed in Table \ref{tab:performance}, our multi-task model requires higher computational resources (262.38 GFLOPs, 130.32 GMACs) compared to PanoFormer (151.63 GFLOPs, 74.85 GMACs) due to its multi-task nature, which simultaneously predicts both depth and surface normals. This increase in computational demand is expected in models that handle multiple tasks, as they inherently require more parameters and operations to integrate and process additional information. Despite the higher complexity, our model maintains a comparable training speed (49 min/epoch) to PanoFormer (47.5 min/epoch) on the 3D60 dataset. The inference time for a single frame differs by less than 10\% (0.1212 sec/instance for our model vs. 0.1109 sec/instance for PanoFormer), indicating that the additional task does not impose a substantial delay in real-time applications.

However, in resource-constrained environments such as mobile devices or embedded systems, where computational power and memory are limited, the model's increased complexity may pose challenges. To mitigate this, we can optimize the model in terms of architecture and training strategies. Architecturally, we can adapt PanoFormer’s block to capture global dependencies on lower-resolution feature maps while using convolutional neural network blocks to extract high-resolution features. This approach reduces computational overhead while preserving the model’s ability to process the complex geometric information in 360° imagery. On the training side, we can implement PyTorch’s mixed precision training, which significantly reduces memory usage and training time. By selectively using full precision where necessary, this method minimizes memory footprint while maintaining high accuracy. It is especially effective on modern hardware, such as NVIDIA GPUs, which have dedicated support for mixed precision operations.

While these optimizations improve the efficiency of our model, additional trade-offs must be considered for deployment in low-resource environments. For example, techniques such as pruning, quantization, or distillation could further reduce resource usage, though they often come at the cost of accuracy. In scenarios where inference speed is critical but high precision is not, these methods may be employed to strike a better balance between performance and resource constraints. Overall, our current implementation demonstrates that, despite its higher complexity, the model remains feasible for environments with moderate computational resources, such as high-end consumer GPUs or cloud-based systems.

\begin{table}[!]
\caption{Performance quantitative comparisons.}
\centering
\resizebox{\linewidth}{!}{
\begin{tabular}{l|c|c|c|c}
\hline
        & GFLOPs  & GMACs  & Training Time & Inference Time \\ \hline
PanoFormer & 151.63 & 74.85  & 47.5 min/epoch   & 0.1109 s  \\ \hline
Ours  & 262.38 & 130.32 & 49 min/epoch   & 0.1212 s \\ \hline
\end{tabular}
}
\label{tab:performance}
\end{table}
\begin{table}[t!]
    \centering
    \caption{Ablation study for individual components.}
		\begin{tabular}  {l|ccc}
			\toprule 
			Method & MAE~$\downarrow$  &ARE~$\downarrow$ &RMSE~$\downarrow$\\
			\midrule
			 Baseline & 0.1577 & 0.0787 & 0.2975 \\
			 Baseline+FB & 0.1295 & 0.0649 & 0.2470 \\
            
			 Baseline+FB+Fusion & 0.1458 & 0.0742 & 0.2682 \\
			 Baseline+FB+Multi-scale & 0.0997 & 0.0480 & 0.2110 \\
			 Ours (all together) & 0.0962 & 0.0465 & 0.2050 \\
			\bottomrule 
	\end{tabular}
    \label{tab:ablation}
\end{table}

\begin{table*}[t!]
    \caption{Quantitative comparisons of different fusion strategies for depth and surface normals on the 3D60 dataset.}
    \centering
        \begin{tabular}{l|cccc|cccc}
            \toprule
            \multirow{2}{*}{Method}
            & \multicolumn{4}{c|}{Depth Estimation} & \multicolumn{4}{c}{Surface Normal Estimation}  \\
            & \raisebox{-0.5ex}{MAE $\downarrow$}  & \raisebox{-0.5ex}{ARE $\downarrow$} & \raisebox{-0.5ex}{RMSE $\downarrow$}  & \raisebox{-0.5ex}{$\delta_{D1}$ $\uparrow$}
            & \raisebox{-0.5ex}{Mean $\downarrow$}  & \raisebox{-0.5ex}{Median $\downarrow$} & \raisebox{-0.5ex}{MSE $\downarrow$}  & \raisebox{-0.5ex}{$\delta_{N1}$ $\uparrow$}\\
            
            \midrule
            \midrule
    
             One encoder&0.0984	&0.0474	&0.2087	&97.59	&5.4010	&0.3594&193.7227&77.92\\
             Cross-attention&0.1026	&0.0496	&0.2141	&97.32	&5.3561	&0.3442&192.6810&78.08\\
             Ours&\textbf{0.0962}	&\textbf{0.0465}	&\textbf{0.2050}	&\textbf{97.66}	&\textbf{5.2394}	&\textbf{0.3025}&\textbf{187.1651}&\textbf{78.19}\\
            \bottomrule
        \end{tabular}
    \label{tab:alternative_archi}
\end{table*}
\subsection{Ablation Study}\label{sec:ablation}
 We conduct an individual component study on the 3D60 dataset to validate the critical components of our multi-task architecture under consistent training and testing conditions, as illustrated in Table~\ref{tab:ablation}. Our baseline model involves duplicating the PanoFormer network for both depth and surface normal tasks, with the two branches only intersecting at the bottleneck block. This configuration has unsatisfactory performance, indicating that a naive combination of networks does not suffice for effective multi-task learning. To evaluate other components in our MTL model, we introduced a shared convolutional feature extraction block (FB) for both branches to extract low-level features. This modification led to a notable improvement of 17.88\% in terms of MAE. 

 Next, we investigate the effectiveness of the fusion blocks and the multi-scale decoder within the existing structure. While these components led to improvements of 7.55\% and 36.78\%, respectively, adding the fusion module alone slightly reduced performance compared to using only the feature embedding block. This is likely due to the increased complexity of shared task information, which the network struggles to process efficiently without the multi-scale decoder to balance and integrate information across different levels. The fusion module is designed to facilitate knowledge transfer between tasks, but without the multi-scale decoder, it cannot fully capitalize on this synergy. However, when all components were integrated into our comprehensive multi-task architecture, performance improved by 39.00\%, demonstrating how the fusion module and multi-scale decoder complement each other to enhance both tasks. This outcome underscores the importance of each component in our multi-task network and highlights how their combined effect is crucial for achieving optimal performance.

\subsection{Alternative Architectures}
We further conduct experiments on two alternative fusion module designs: (1) using a shared encoder for both depth and surface normal estimation, and (2) applying a cross-domain attention mechanism in the decoder to bridge the gap between the two tasks. Our proposed fusion module still outperforms these alternatives in both depth and surface normal estimation, as detailed in Table~\ref{tab:alternative_archi}.

\subsubsection{Shared Encoder for Both Tasks}
In the first experiment, we used a single shared encoder for both tasks. This approach (One encoder) has the advantage of reducing the overall number of parameters and simplifying the model architecture. However, while this strategy achieved acceptable results, especially in the depth estimation task, it could not fully differentiate between the specific characteristics of depth and surface normal features. Each task requires different feature representations, particularly in the early stages of encoding, where distinct geometric cues are crucial for accurate predictions. As a result, the shared encoder was unable to capture these nuances effectively, leading to suboptimal performance in surface normal estimation. The error metrics show that this approach led to an imbalance between the tasks, with the depth task dominating the other during training, which limited the model’s ability to generalize well across both tasks simultaneously.

\subsubsection{Cross-Domain Attention in the Decoder}
The second experiment involves using a cross-domain attention mechanism to bridge depth and surface normal estimation in the decoder stage. While this design (Cross-attention) is intended to enhance information exchange between the two tasks during decoding, it introduces excessive complexity to the model. The resulting architecture becomes too large to train at full scale, significantly increasing computational demands and memory requirements. As a result, we did not apply cross-attention at the final scale. The results in Table~\ref{tab:alternative_archi} show that this strategy caused one task to dominate the other during training. Surface normal estimation achieved better quantitative results than depth estimation, indicating an imbalance in task performance.

\subsubsection{Our Fusion Module}
In contrast, our proposed fusion module achieves an appropriate balance between the two tasks by facilitating smooth information transfer without overwhelming the network with excessive complexity. The separate encoders allow for task-specific feature extraction, and the fusion module effectively shares useful information between the two branches without leading to overfitting or imbalance in the training for each task. As a result, both tasks achieve their best results simultaneously. This demonstrates the effectiveness of our fusion module and the entire architecture in maintaining high performance across both depth and surface normal estimation.

\begin{figure*}[t]
    \centering
    \includegraphics[width=\linewidth]{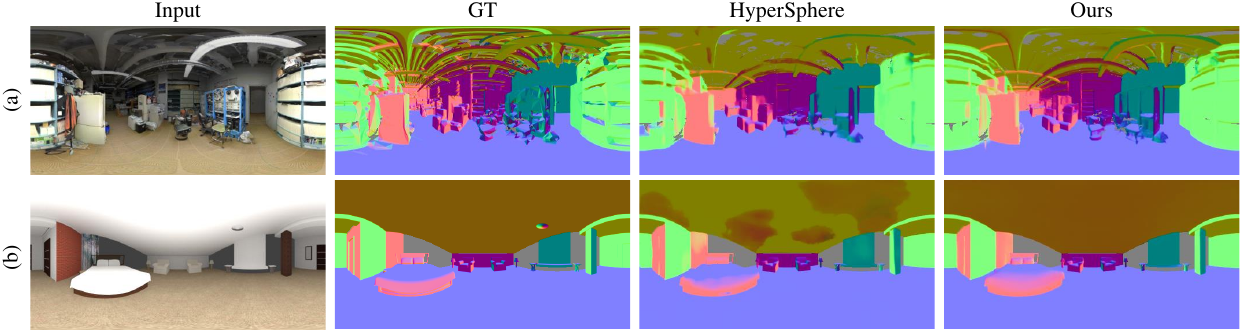}
    \caption{
    Qualitative evaluation on two specific examples from the Stanford2D3D (a) and SunCG (b) datasets. Our model demonstrates more precise predictions, accurately capturing object boundaries and the entire ceiling.
    }
    \label{fig:limitation}
\end{figure*}

\begin{figure*}[!]
    \centering
    \includegraphics[width=\textwidth]{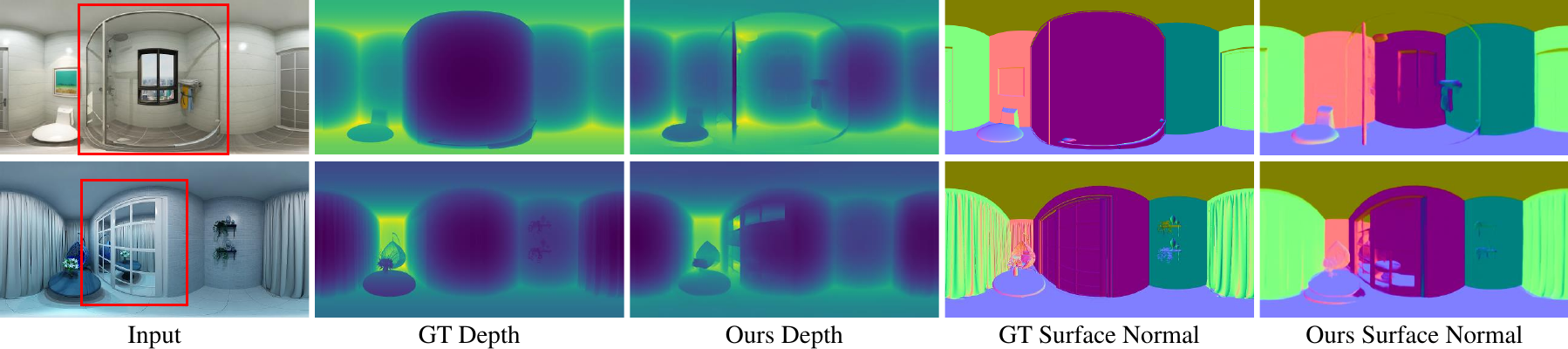}
    \caption{
    Failure cases involving glass walls in the top row examples and mirror scenes in the bottom row from the Structured3D dataset. The red rectangle indicates such regions.
    }
    \label{fig:failure_case}
\end{figure*}

\subsection{Limitations and Future Works}\label{sec:limitation}
We evaluate the performance of 360° depth and surface normal tasks, where our model sets a new state-of-the-art performance in depth estimation across all metrics, and in surface normal estimation across most metrics except for the median error, as shown in Table~\ref{tab:normal_quantitative_comparison}. For 360° surface normal estimation, we primarily compare our model with the current state-of-the-art, HyperSphere, and consistently observe lower Mean and MSE metrics, though with a higher Median error. Individual quantitative results from the Stanford2D3D (example (a)) and SunCG (example (b)) datasets are detailed in Table~\ref{tab:limitation}. The low Mean error values reflect that, on average, our model's predictions closely align with true surface orientations compared to HyperSphere (3.1416 versus 6.4296 in example (b)). The low MSE, which penalizes larger errors more heavily, further suggests that our model avoids significant outliers, ensuring stable and accurate predictions across most of the image (601.8008 versus 715.2992 in example (a)). 

However, the higher Median error indicates challenges with specific outliers or complex regions (5.0646 versus 2.6880 in example (a), and 0.4131 versus 0.2639 in example (b)). We visualize these examples in Fig.~\ref{fig:limitation}, where our method captures more geometric details around small object boundaries in (a) and accurately predicts the ceiling's surface normals, whereas HyperSphere struggles, in (b). While the consistently low Mean and MSE errors highlight our model's effectiveness in delivering accurate surface normal predictions across a wide range of scenarios, the higher Median error points to areas for future improvement. Specifically, addressing the occasional difficulties our model faces with outliers and certain challenging regions will be a key focus in our future work. 

\begin{table}[!]
    \centering
    \caption{Examples from (a) Stanford2D3D and (b) SunCG datasets.}
    \resizebox{\linewidth}{!}{
    \begin{tabular}  {c|c|rcc}
			\toprule 
			Example & Method & Mean~$\downarrow$  &Median~$\downarrow$ &MSE~$\downarrow$ \\
			\midrule
              \multirow{2}{*}{\centering (a)}&
			 HyperSphere & 15.1213 & 2.6880 & 715.2992 \\    
			 & Ours & 14.3891 & 5.0646 & 601.8008 \\
		   \midrule
              \multirow{2}{*}{\centering (b)}&
			 HyperSphere & 6.4296 & 0.2639 & 146.5674 \\    
			 & Ours & 3.1416 & 0.4131 & 75.3159 \\
			\bottomrule 
	\end{tabular}
    }
    \label{tab:limitation}
\end{table}

Additionally, our model encounters issues when dealing with scenes containing glass walls or mirrors, as illustrated in Fig.~\ref{fig:failure_case}. In such scenarios, the model often struggles to accurately interpret surface orientations and depth due to the unique visual effects introduced by these materials. For instance, mirrors can reflect entire sections of a room, causing the model to perceive over-extended scene depth and misinterpret surface normals. This occurs because the model treats the reflection as a continuation of the physical environment, resulting in erroneous geometric predictions. Similarly, when transparent glass is present in front of the camera, the model may over-extrapolate depth or surface normal values by mistakenly interpreting the space behind the glass as part of the scene geometry, despite the distortion caused by the transparency.

To address these limitations, future research could explore strategies to handle reflective and transparent surfaces. One potential solution is incorporating additional material-based cues that allow the model to differentiate between glass, mirrors, and solid objects. This could involve using reflectance maps or leveraging external sensors that detect surface properties beyond visual data. Another approach could include semantic segmentation in the multi-task framework, enabling the model to identify and treat reflective or transparent objects differently during depth and surface normal estimation. These enhancements will make the model more robust for practical applications, such as indoor navigation and scene reconstruction, where these materials are common.

\section{Conclusion}\label{sec:conclusion}
In this paper, we propose an MTL network for monocular indoor 360° geometric estimation, achieving state-of-the-art performance in both depth and surface normal tasks simultaneously. Our architecture leverages the strengths of MTL to provide a comprehensive understanding of scene geometry by effectively fusing features from both depth and surface normal estimations. We introduce a fusion module composed of specifically designed blocks to facilitate positive knowledge transfer between the two ViT branches. Additionally, our multi-scale spherical decoder further enhances the perception of scene structure at various levels. Experimental results demonstrate that our approach establishes new baselines in both tasks, highlighting our model's superior performance, robustness and generalizability. This is further evidenced by conducting experiments on the Structured3D dataset, underscoring its potential applicability in real-world scenarios.

\section*{Declarations}
\subsection*{Availability of Data and Materials}
The training and testing datasets are publicly available online.
\subsection*{Competing interests}
The authors have no competing interests to declare that are relevant to the content of this article.
\subsection*{Funding}
This research was supported by the Marsden Fund Council managed by the Royal Society of New Zealand under Grant No. MFP-20-VUW-180 and the Royal Society (UK) under Grant No. IES\textbackslash R1\textbackslash 180126.
\subsection*{Authors' Contributions}
K. Huang designed and developed the deep architecture, conducted the experiments, and wrote the paper.
K. Huang and F.-L. Zhang initiated the project and the collaboration.
F.-L. Zhang contributed to the method design, experiment design and paper editing.
F. Zhang contributed to the method design and paper editing.
Y.-K. Lai and P. Rosin contributed to the method design and experiment design. 
N.A. Dodgson contributed to the method design and paper editing. 
\subsection*{Acknowledgements}
We thank the funding agencies for supporting this work, as detailed in the Funding section above.
\subsection*{Authors' Information}
Please refer to Author biography section

\bibliographystyle{CVMbib}
\bibliography{Main_file}

\subsection*{Author biography}
\begin{biography}[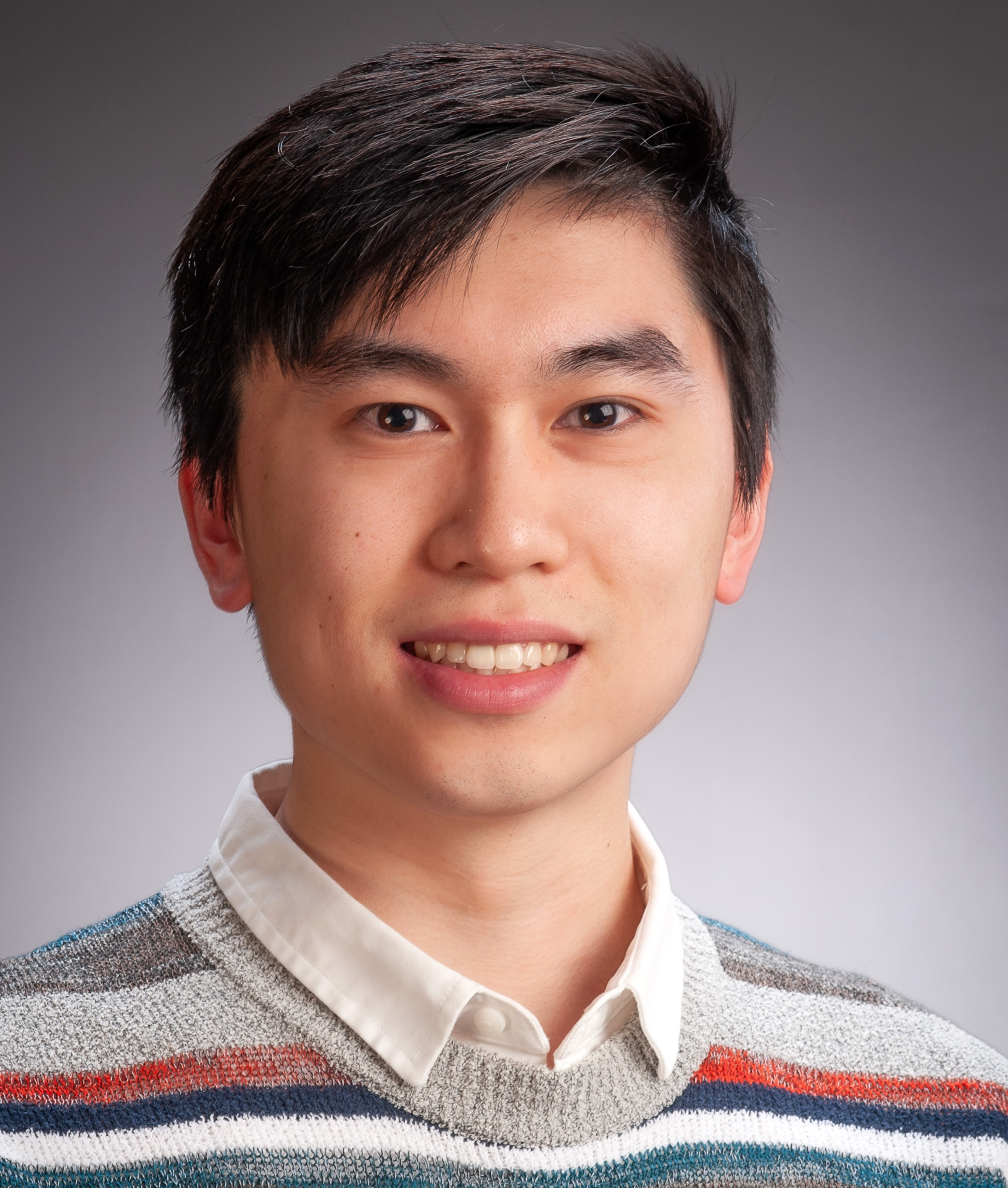]{Kun Huang} is currently a PhD candidate at Victoria University of Wellington, New Zealand. He received a Bachelor's and M.S. degree from Victoria University of Wellington in 2017 and 2021, respectively. His research interests include 360° image and video editing, computer vision, virtual reality and mixed reality. He is a student branch chair and professional activity coordinator of the IEEE New Zealand Central Section.
\end{biography}

\vspace*{1.6em}
\begin{biography}[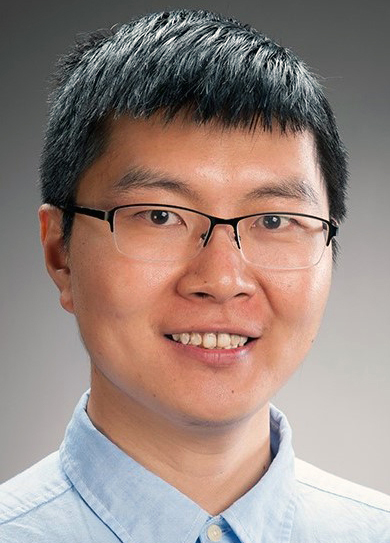]{Fang-Lue Zhang}
received the Ph.D. degree from Tsinghua University, in 2015. He is currently a Senior Lecturer in Computer Graphics at the Victoria University of Wellington, Wellington, New Zealand. His research interests include image and video editing, computer vision, and computer graphics. He received the Victoria Early-Career Research Excellence Award, in 2019, and the Fast-Start Marsden Grant from the New Zealand Royal Society, in 2020. He is on the editorial board of Computer \& Graphics. He served as program chair of Pacific Graphics 2020 \& 2021, and CVM 2024. He is a member of ACM and a committee member of IEEE Central New Zealand Sector. 
\end{biography}

\vspace*{1.6em}
\begin{biography}[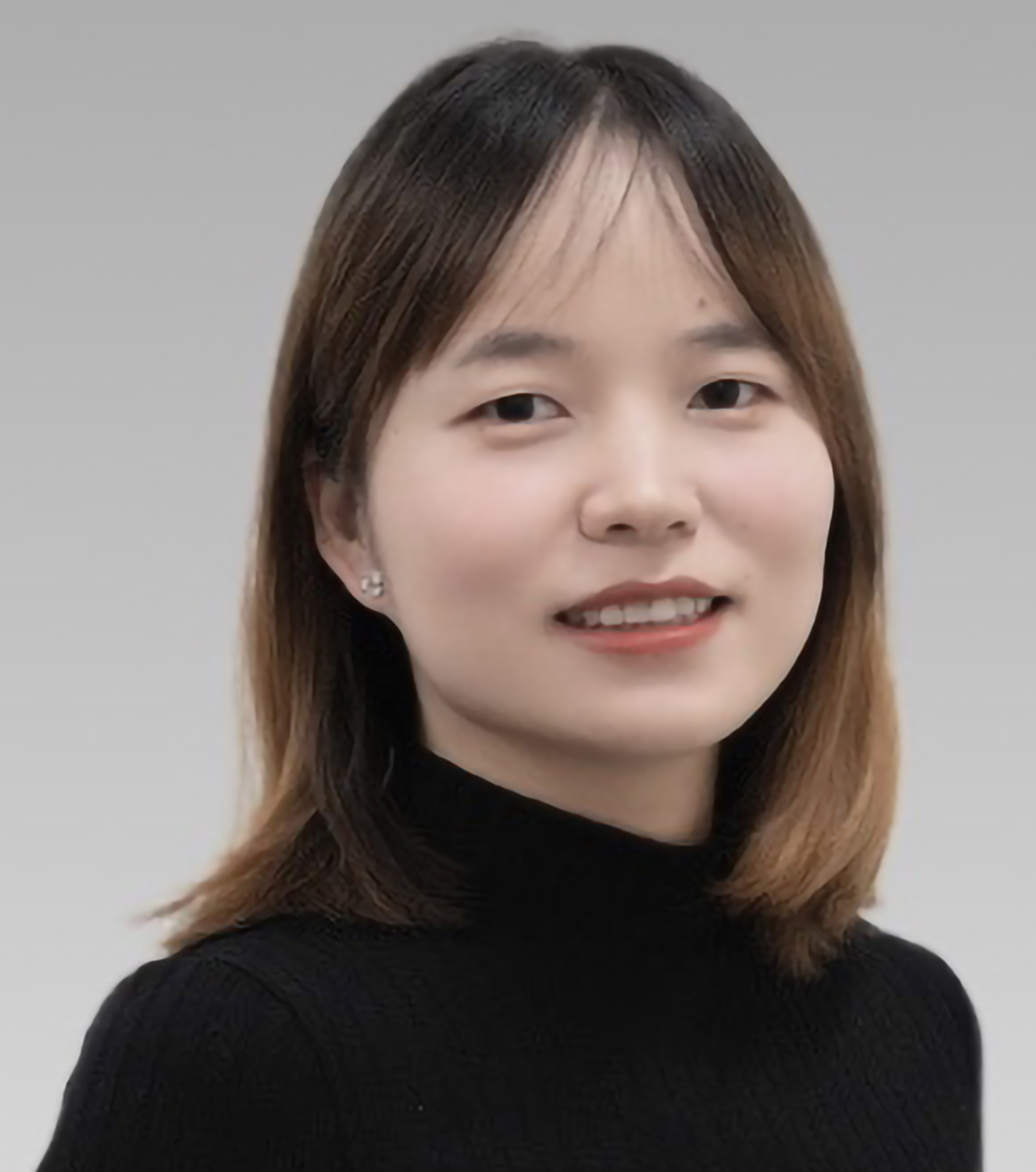]{Fangfang Zhang} received the
Ph.D. degree in computer science from Victoria
University of Wellington, New Zealand, in 2021. Her PhD thesis received the ACM SIGEVO Dissertation Award, Honorable Mention, and IEEE CIS Outstanding PhD Dissertation Award. She is currently a
lecturer with the Centre for Data Science and Artificial
Intelligence \& School of Engineering and
Computer Science, Victoria University of Wellington,
New Zealand. She has over 65 papers in refereed
international journals and conferences. Her research
interests include evolutionary computation, hyperheuristic
learning/optimisation, job shop scheduling, surrogate, and multitask
learning. Dr. Fangfang is an Associate Editor of Expert Systems With
Applications, and Swarm and Evolutionary Computation. 
She is the secretary of the IEEE New Zealand Central Section.
She is a Vice-Chair of the
IEEE Taskforce on Evolutionary Scheduling and Combinatorial Optimisation.
\end{biography}

\vspace*{1.6em}
\begin{biography}[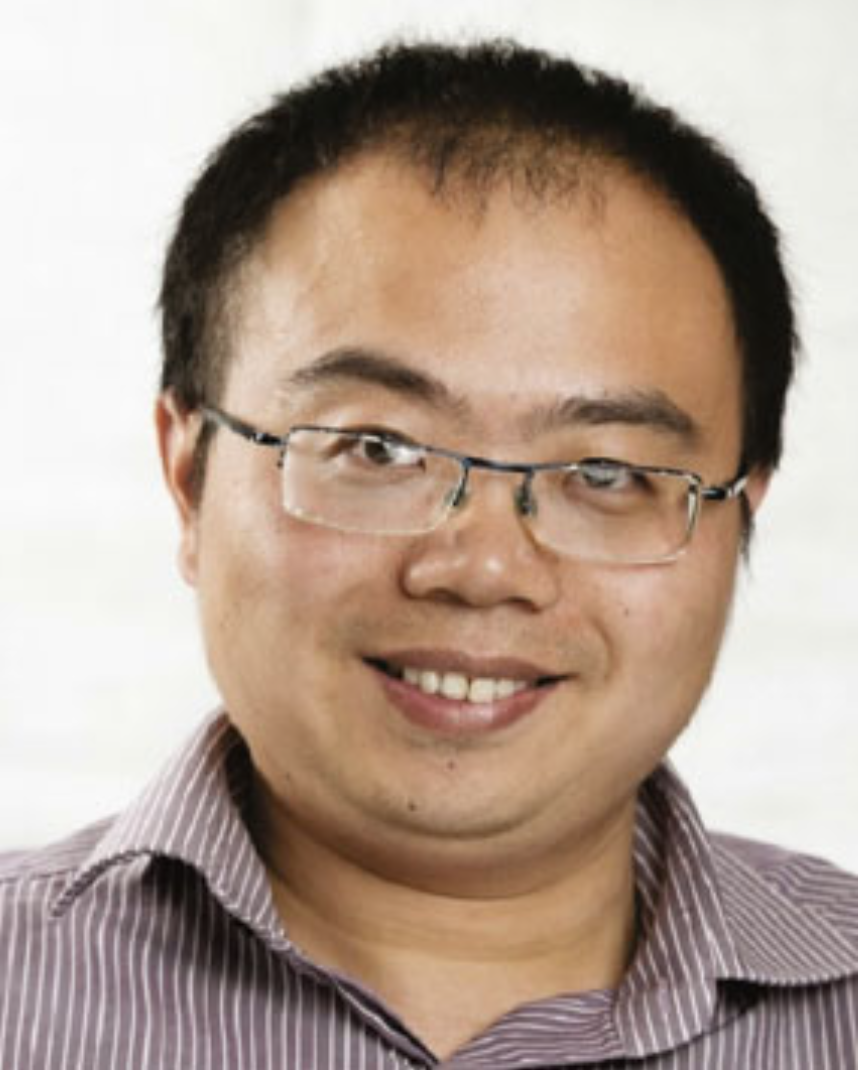]{Yu-Kun Lai}
Yu-Kun Lai is a professor in the School of Computer Science and Informatics, Cardiff University. He received his B.S. and Ph.D. degrees in computer science from Tsinghua University, China, in 2003 and 2008 respectively. His research interests include computer graphics, computer vision, geometric modeling,
and image processing.
\end{biography}

\vspace*{1.6em}
\begin{biography}[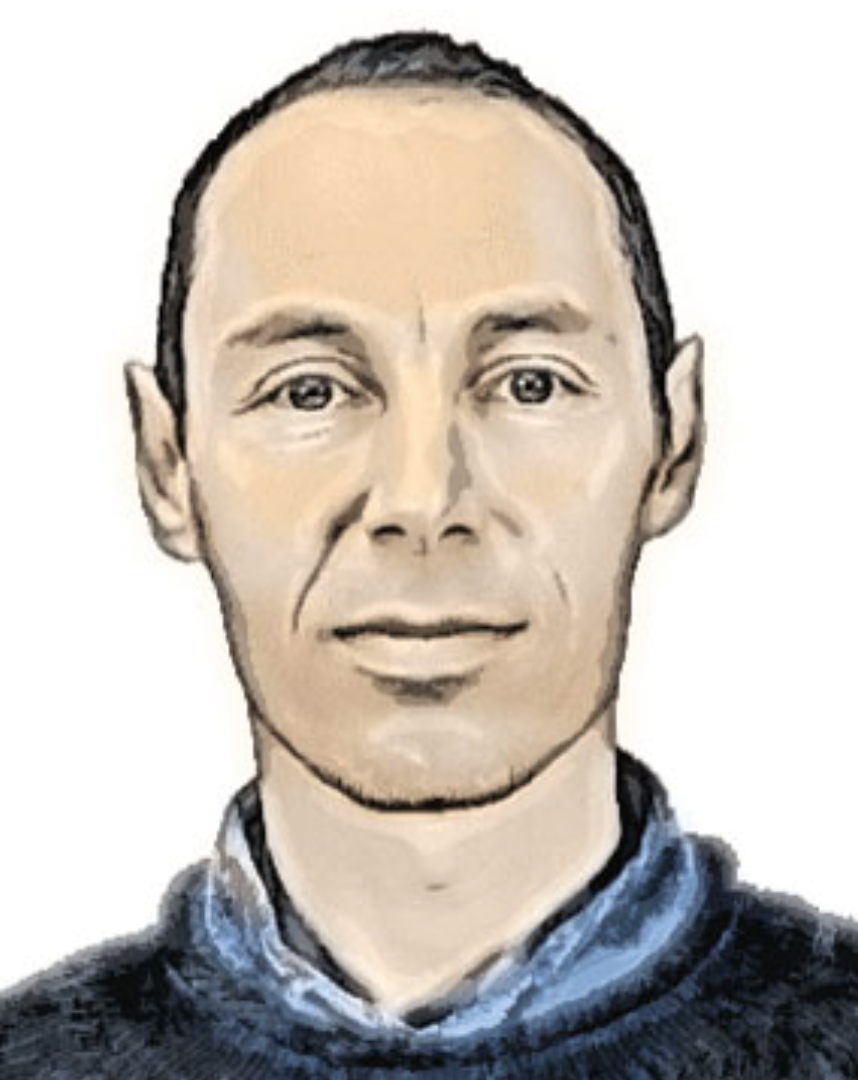]{Paul L. Rosin}
Paul L. Rosin is a professor in the School of Computer Science and Informatics, Cardiff University, UK. He received his Ph.D. degree from City University, London, in 1988. Previous posts were at Brunel University, UK; the Institute for Remote Sensing Applications, Joint Research Centre,
Italy; and Curtin University of Technology, Australia. His research interests include low-level image processing, performance evaluation, shape analysis, facial analysis, medical image analysis, 3D mesh processing, cellular automata, non-photorealistic rendering, and cultural heritage.
\end{biography}

\vspace*{1.6em}
\begin{biography}[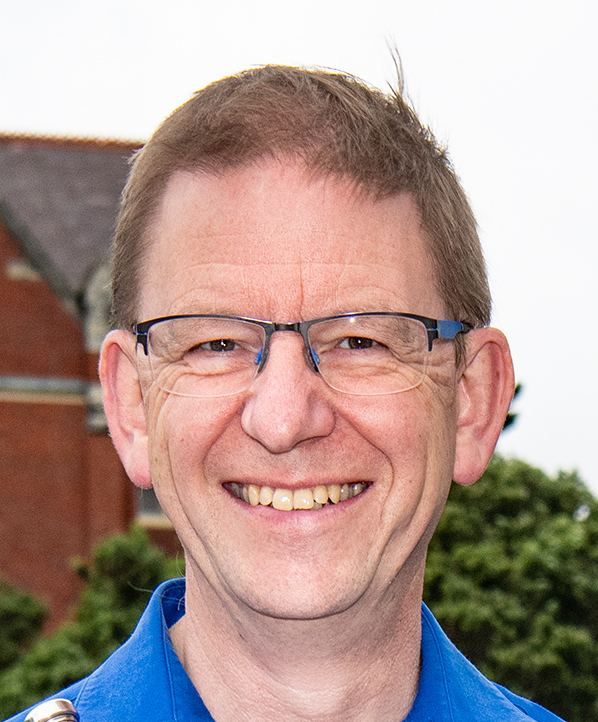]{Neil Dodgson}
Neil Dodgson is Professor of Computer Graphics and Dean of Graduate Research at Victoria University of Wellington. His PhD is in image processing, from the University of Cambridge. He spent 25 years at Cambridge, becoming full professor in 2010. He moved to Wellington in 2016 to lead the computer graphics group there. His research is in 3D TV, subdivision surfaces, imaging, and aesthetics. He is a Chartered Engineer and a Fellow of Engineering New Zealand and of the Institution of Engineering and Technology (IET) and the Institute for Mathematics and its Applications (IMA) in the UK.
\end{biography}

\vspace*{2.6em}

\end{document}